\documentclass[pdflatex,sn-mathphys-num]{sn-jnl}% Math and Physical Sciences Numbered Reference Style
\usepackage{graphicx}%
\usepackage{multirow}%
\usepackage{amsmath,amssymb,amsfonts}%
\usepackage{amsthm}%
\usepackage{mathrsfs}%
\usepackage[title]{appendix}%
\usepackage{xcolor}%
\usepackage[table]{xcolor}
\usepackage{textcomp}%
\usepackage{manyfoot}%
\usepackage{booktabs}%
\usepackage{algorithm}%
\usepackage{algorithmicx}%
\usepackage{algpseudocode}%
\usepackage{listings}%
\usepackage{comment}
\usepackage{caption}%
\usepackage{subcaption}%
\usepackage{multicol}%
\usepackage{tabularx}%
\usepackage{longtable}%
\usepackage{adjustbox}%
\usepackage{wrapfig}%
\usepackage{makecell}%
\usepackage{esvect}%
\usepackage{hyperref}%
\usepackage{pdflscape}%
\usepackage{afterpage}%
\usepackage[strings]{underscore}
\usepackage{anyfontsize}

\theoremstyle{thmstyleone}%
\theoremstyle{thmstyletwo}%
\usepackage{tikz}
\usetikzlibrary{positioning}
\usetikzlibrary{arrows.meta}
\theoremstyle{thmstylethree}%
\newtheorem{definition}{Definition}%

\definecolor{blue_comment}{rgb}{0.2, 0.4, 0.7}
\newcommand{\oana}[1]{\textcolor{blue_comment}{\textbf{Oana:} #1}}
\definecolor{green_comment}{rgb}{0.3, 0.5, 0.3}

\definecolor{red_comment}{rgb}{0.7, 0.3, 0.4}

\newcommand{\task}[1]{\paragraph{Task Description}  #1}
\newcommand{\zeroshot}[1]{\paragraph{Annotation-free}  #1}
\newcommand{\datasets}[1]{\paragraph{Annotated datasets}  #1}
\newcommand{\solutions}[1]{\paragraph{Models}  #1}

\begin{document}

\title[Article Title]{Detecting Greenwashing: A Natural Language Processing Literature Survey}
% Greenwashing Detection: NLP Techniques, Datasets, and Evaluation


\author*[1,2,3,4]{\fnm{Tom} \sur{Calamai}}\email{tom.calamai@telecom-paris.fr}

\author[1,2]{\fnm{Oana} \sur{Balalau}}\email{oana.balalau@inria.fr}

\author[4]{\fnm{Théo} \sur{Le Guenedal}}\email{theo.leguenedal-ext@amundi.com}
\author[2,3]{\fnm{Fabian} \sur{Suchanek}}\email{suchanek@telecom-paris.fr}

\affil[1]{ \orgname{LIX, CNRS, Inria, École polytechnique}, \orgaddress{\country{France}}}
\affil[2]{ \orgname{Institut Polytechnique de Paris}, \orgaddress{ \country{France}}}
\affil[3]{\orgdiv{DIG}, \orgname{Télécom Paris}, \orgaddress{ \country{France}}}

\affil[4]{\orgdiv{Innovation Lab}, \orgname{Amundi Technology}, \orgaddress{\country{France}}}

% \affil[3]{\orgdiv{Department}, \orgname{Organization}, \orgaddress{\street{Street}, \city{City}, \postcode{610101}, \state{State}, \country{Country}}}


\abstract{
%Please provide an abstract of 150 to 250 words. The abstract should not contain any undefined abbreviations or unspecified references.

Greenwashing refers to practices by corporations or governments that intentionally mislead the public about their environmental impact. This paper provides a comprehensive and methodologically grounded survey of natural language processing (NLP) approaches for detecting greenwashing in textual data, with a focus on corporate climate communication.  

Rather than treating greenwashing as a single, monolithic task, we examine the set of NLP problems, also known as climate NLP tasks, that researchers have used to approximate it, ranging from climate topic detection to the identification of deceptive communication patterns. Our focus is on the methodological foundations of these approaches: how tasks are formulated, how datasets are constructed, and how model evaluation influences reliability.

Our review reveals a fragmented landscape: several subtasks now exhibit near-perfect performance under controlled settings, yet tasks involving ambiguity, subjectivity, or reasoning remain challenging.  Crucially, no dataset of verified greenwashing cases currently exist.

We argue that advancing automated greenwashing detection requires principled NLP methodologies that combine reliable data annotations with interpretable model design. Future work should leverage third-party judgments, such as verified media reports or regulatory records, to mitigate annotation subjectivity and legal risk, and adopt decomposed pipelines that support human oversight, traceable reasoning, and efficient model design. }

\keywords{Greenwashing detection, Natural Language Processing, Climate NLP, Climate communication, Corporate sustainability reporting}



\maketitle

\section{Introduction}
\label{sec:intro}

%%%%%%%%%%%%% Climate change is important

Since the first report of the Intergovernmental Panel on Climate Change in 1990 \citep{change1990ipcc},
world leaders and citizens have increasingly recognized the significant adverse impacts of human activities on the environment, particularly on climate change  \citep{climatebriefhistory}. This increased awareness has translated into laws and investments, such as the European Green Deal~\citep{europeangreendeal} or the Inflation Reduction Act in the US  \citep{inflationact}. Many companies have used the financial incentives offered by states, and the guidelines and legislation, to make significant steps towards sustainability  \citep{greencompanies}. At the same time, this growing attention also generated an advertising opportunity for companies that aim to promote themselves as environmentally aware and responsible. 
Indeed, some companies have been found to deliberately manipulate their data and statistics to appear more environment-friendly. The Diesel Scandal around the Volkswagen car company is a prominent example  \citep{volkswagen}. However, such cases are not the norm. More commonly, companies avoid outright data manipulation but present themselves in a misleadingly positive light regarding their environmental impact, a practice called \emph{greenwashing}. \smallskip

Greenwashing is a widely observed practice rather than a marginal occurrence: the Australian Competition \& Consumer Commission performed a sweep analysis on 247 businesses in October 2022 and found that $57\%$ of them made misleading claims about their environmental performance  \citep{sweepaustralia}. The International Consumer Protection Enforcement Network, too, found in November 2020 that $40\%$ of around 500 websites of businesses covering different sectors contain misleading green claims  \citep{globalsweep}.\smallskip

Addressing greenwashing is not only in the broader public interest but also in the company's best interest, as the practice is often subject to legal sanctions prohibiting misleading or deceptive conduct, with businesses facing fines or lawsuits as a consequence  \citep{litigatorsgreenwashing}. In addition, political awareness and public awareness of the issue via traditional and social media  \citep{bonpote} can harm the reputation of a company involved in the practice. These factors make greenwashing a risk factor that can affect investments and that should be monitored at the company and portfolio levels.\smallskip

The growing public and regulatory attention to corporate climate claims raises a central question: \textbf{to what extent can greenwashing be detected automatically?} Automated detection systems could support investors, journalists, prosecutors, and activists in identifying potentially misleading or illicit communication, while also helping companies avoid unintentionally deceptive disclosures. Yet despite the relevance of this question, the methodological foundations of NLP-based greenwashing detection remain fragmented.

\paragraph{Methodological grounding} In this paper, we provide \textbf{the first comprehensive and methodologically grounded survey of NLP techniques for identifying potentially misleading climate-related corporate communication}. We analyze \textbf{61 scientific works} that examine one or more aspects of greenwashing, ranging from identifying climate-related statements and assessing environmental claims to evaluating specificity, stance, argumentation, and the presence of deceptive linguistic cues. Unlike prior surveys \citep{moodaley_greenwashing_2023, measuring_greenwashing}, which primarily summarize findings reported in the literature, our focus is on the \textbf{methodological foundations} of these approaches: \textbf{how tasks are formulated, how datasets are constructed, and how model evaluation influences reliability}.

\paragraph{Summary of findings}
Recent advances in NLP have enabled increasingly sophisticated analyses of climate-related corporate communication, yet our survey shows that progress across the building blocks of greenwashing detection is highly uneven. Several tasks, such as climate topic identification, climate risk classification, and the characterization of environmental claim types, now achieve near perfect performance when tackled with fine-tuned models. Even simple keyword-based approaches often achieve surprisingly strong results, suggesting that many of these tasks are driven by distinctive lexical terms rather than deep domain knowledge. 

In contrast, tasks that require nuance, subjectivity, or reasoning, such as distinguishing specific from vague commitments or assessing rhetorical deception, remain challenging. Low inter-annotator agreement, ambiguous label definitions, and small datasets, undermine the reliability of the reported performance. This pattern reflects a broader issue: many real indicators of greenwashing require judgment on subtle differences, multi-document context, or implicit reasoning, where current models struggle and where evaluation is especially difficult.

These challenges are rendered more difficult by methodological weaknesses in research papers that limit the interpretability and reproducibility of results. Evaluation practices are often inadequate: few studies report uncertainty estimates, many rely on accuracy without considering semantic proximity between labels, and baselines are frequently missing. Only a minority compare to random, majority, or keyword-based baselines, even though these can drastically change how model performance is interpreted. Furthermore, many experiments are conducted on curated, balanced, and noise-free datasets, conditions that that might not hold in real-world datasets.

A more fundamental limitation is the \textbf{absence of datasets containing verified instances of greenwashing}. While the literature proposes theoretical indicators such as overly positive sentiment, rhetorical cues, and stance inconsistencies, none are validated against real cases. As a result, most studies evaluate proxy tasks rather than greenwashing, leaving the field without empirical foundations for end-to-end detection. 

We believe that creating a greenwashing dataset requires  principled data collection grounded in third-party judgments (e.g., regulators, courts, investigative journalism), clearer and narrower definitions of misleading communication, and decomposed pipelines that combine automated analysis with human oversight.

In conclusion, these findings show that while some subtasks appear solved, building a reliable and trustworthy NLP system for detecting greenwashing remains an open challenge. 

%True progress will require innovative modeling approaches, rigorous evaluation standards, transparent data, and close alignment with legal and regulatory definitions of greenwashing.

\paragraph{Structure of the survey}
The remainder of the paper proceeds as follows. We begin by situating our work within the broader literature on greenwashing and automated climate communication analysis (Section~\ref{sec:related}). We then outline the regulatory landscape that shapes how greenwashing is defined (Section~\ref{sec:definitions}), as these definitions constrain what can be annotated or modeled.
Section~\ref{sec:intermediary tasks} introduces the building-block tasks that underpin automated greenwashing detection: 
identifying climate-related statements; identifying topics aligned with frameworks such as the Task Force on Climate-related Financial Disclosures (TCFD) or Environmental, Social, and Governance (ESG) factors; assessing the presence of green claims; and examining sentiment, argumentation quality, deceptive practices, and stance to discern the tone and authenticity of a message. For each task, we give the \textbf{task description, annotated or annotation-free datasets, models and an analysis of their performance, and finally insights,} in which we highlight important limitations and potential future direction.  
Section~\ref{sec: greenwashing signals} reviews approaches that aim to identify greenwashing as a whole. Finally, Sections~\ref{sec:challenges} and~\ref{sec:conclusion} discuss open methodological challenges and outline research opportunities for building principled, reliable systems for greenwashing detecting.

\section{Related Work} 
\label{sec:related}

\citet{moodaley_greenwashing_2023} analyze the literature on greenwashing, sustainability reporting, and the intersection with research in Artificial Intelligence (AI) and Machine Learning (ML). The authors provide an overview of how the literature has changed over time by examining the number of publications and the different keywords covered by such publications. Unlike our work, this survey does not describe the actual approaches of the publications.
\citet{measuring_greenwashing} review the definitions of greenwashing in business, management, and accounting. They conclude that there is no agreed-upon definition of greenwashing, except that it is defined as a gap between the company's actions and its disclosure. In contrast to our study, the review provides only a brief overview of the techniques for detecting greenwashing.
While our review focuses on the semantic and linguistic aspects of greenwashing identification, other works focus on quantitative approaches that can serve the same objective \citet{dao2024introduction}. \citet{LUBLOY2025123399} studied the quantification of firm-level greenwashing; however, our work is focused on natural language processing approaches, while theirs is more generic.\smallskip

%https://www.researchgate.net/publication/389276920_Mapping_the_greenwashing_research_landscape_a_theoretical_and_field_analysis#fullTextFileContent, a greenwashing survey on the research landscape but very generic

Other works are less directly related to our survey. For example, a growing community is focused on using AI to tackle climate change\footnote{See for example \url{https://www.climatechange.ai/}}. \citet{Rolnick_tackling_climate} review how machine learning has been used and could be used to tackle climate change. They explore different sectors of the economy, such as electricity, transportation, industry, and agriculture, among others. The survey discusses climate investments and the effects of climate change on finance, but not greenwashing. 
\citet{alonsoMachineLearningMethods2023} provide a survey on ML approaches for climate finance, also known in the literature as green finance or carbon finance. Climate finance is a subfield in economics dedicated to providing tools from financial economics to tackle climate change. Finally, several works study the impact of AI on climate change  \citep{hershcovich_towards_2022,kaack_aligning_2022,rohde_broadening_2023,verdecchia_systematic_2023}. Our work, in contrast, surveys approaches that use AI to detect greenwashing.

\section{Preliminaries}
\label{sec:definitions}

\subsection{Legal Context}
In this section, we provide the legal context for the definition of greenwashing. 
We first present international and national laws, followed by how they are applied in the corporate context.

\paragraph{International and Nation Level Legislation and Guidelines on Climate Change Mitigation}
The first international summit on the effect of humans on the environment was the United Nations Conference on the Human Environment, held in 1972 in Sweden \citep{timelineclimate}. The \textit{Intergovernmental Panel on Climate Change (IPCC)} was created in 1988 and released its first report in 1990  \citep{change1990ipcc}. It was only in 1992 that the first treaty between countries was signed, the United Nations Framework Convention on Climate Change (UNFCCC) \citep{timelineclimate}. The treaty only encouraged countries to reduce their emissions, and in 1997, the Kyoto Protocol set commitments for the countries to follow \citep{kyotoprotocol}. The 36 countries that participated in the Protocol reduced their emissions in 2008–2012 by a large margin in respect to the levels in 1990; however, the global emissions increased by $32\%$ in the same period. 
Since the creation of UNFCCC, the nations, i.e., the parties, have met yearly at the ``Conference of the Parties'' (\textit{COP}). \smallskip

The three main directions of \textit{climate change policy} are reducing greenhouse emissions, promoting renewable energy, and improving energy efficiency. One major milestone was the introduction in 2005 of the European Union Emissions Trading System, the world's first large-scale emissions trading scheme. The trading system limits the amount of CO2 emitted by European industries and covers $46\%$ of the EU's CO2 emissions. It is estimated that the trading system reduced CO2 emissions in the EU by $3.8\%$ between 2008 and 2016 \citep{bayer2020european}. At COP21, in 2015, 194 nations plus the European Union signed the Paris Agreement, a treaty by which the nations commit to keep the rise in global surface temperature below 2 °C (3.6 °F) above pre-industrial levels. To achieve this goal, each country sends a national climate action plan every five years, and the parties assess the collective progress made towards achieving the climate goals. 
The first such assessment took place at COP28, where it was established that an important direction to achieve the long-term goals of countries was to transition away from fossil fuels to renewable energy. In 2021, the European Climate Law was adopted: the EU commits to reducing its emissions by at least $55\%$ by 2030 with respect to the 1990 levels and becoming climate neutral by 2050 \citep{euclimatelaw}. 
Similarly, countries such as Canada \citep{canadaclimatelaw}, Taiwan \citep{taiwanclimatelaw}, South Korea \citep{koreaclimatelaw}, and Australia \citep{australiaclimatelaw}, among others, aim to achieve carbon neutrality by 2050 and have passed laws setting this goal.\smallskip

\paragraph{Corporate Level Laws and Guidelines} Laws and regulations at international and national levels have repercussions on companies that must comply or risk fines. However, assessing if a company has taken the necessary steps is not trivial, given that some laws refer to how a company will operate in the future, for example, by polluting less. Even without laws, investors can be concerned about environmental or social issues; hence, companies have used voluntary disclosures for a long time.  \smallskip

One of the first organizations to provide standards for reporting on climate change or social aspects is the  Global Reporting Initiative (GRI), with its first guidelines published in 2000 \citep{grihistory}.
In 2015, the Financial Stability Board and the Group of 20 created the \textit{Task Force for Climate-related Financial Disclosure (TCFD)} guidelines on disclosure in response to shortcomings of COP21, in particular the lack of standards climate-related disclosure \citep{enwiki:1257716178}.
In 2021, the International Sustainability Standards Board (ISSB) was created to establish standards for climate-related disclosure, and starting in 2024, the standard released by this board will be applied worldwide \citep{ifrs}. ISSB standards were aligned with GRI disclosure standards to make them complementary and interoperable \citep{gri}, while the ISSB standards are expected to take over TCFD \citep{ifrs}.
Differently from the voluntary reporting standards of GRI, TCFD, and ISSB, the Corporate sustainability reporting law in the EU required the creation of the European Sustainability Reporting Standards (ESRS), which are mandatory for companies subject to EU law \citep{esrs}. ESRS has high interoperability with GRI and ISSB. GRI, TCFD, ESRS and ISSB fall under the umbrella term of environmental, social, and governance (ESG) guidelines for company disclosure. 
The European Union created its first law obliging companies to provide non-financial disclosure reporting in 2014, the Non-Financial Reporting Directive, which focused disclosure on environmental and social aspects. In 2023, the EU expanded this legislation via the Corporate Sustainability Reporting Law. Some European countries anticipated this legislation with their own, for example, the 2001 New Economic Regulations Act in France. Switzerland, which is not part of the EU, has imposed a mandatory TCFD disclosure for large public companies, banks, and insurance companies starting from January 2024 \citep{disclosureswiss}, and similar laws exist in New Zealand \citep{disclosurezealand}. While not compulsory, initiatives like the Carbon Disclosure Project (CDP) have played a pivotal role by standardizing responses related to the climate disclosure of a company via structured questionnaires, thereby facilitating more systematic and comparable reporting.  

\subsection{Definition of Greenwashing}

A widely cited and comprehensive definition, synthesizing those commonly found in the literature, is provided by the Oxford English Dictionary \citep{GreenwashMeaningsEtymology2023}.

\begin{definition}\textbf{Greenwashing:}
    \label{def:greenwashing}
    To mislead the public (or to counter public or media concerns) by falsely representing a person, company, product, etc., as environmentally responsible.
\end{definition}

While individuals, companies, or countries can all engage in greenwashing, we will focus on climate-related greenwashing by companies in this survey, with the following definition: 
\begin{definition}
\label{def:greenwashing2}
   \textbf{Corporate climate-related greenwashing:} To mislead the public into falsely representing the effort made by a company to achieve its carbon transition.
\end{definition}

As climate-related disclosures face increasing regulation and greenwashing poses significant risks to a company's reputation, some companies adopt a strategy of silence, avoiding discussions about their environmental impact. This deliberate lack of communication is known as \textit{greenhushing} \citep{Letzing}. These definitions imply that greenwashing is a deliberate act; however, in many cases, it results from an error or miscommunication by companies genuinely trying to showcase their sustainability efforts, and that are trying to best follow disclosure standards.\smallskip

As shown by Definitions \ref{def:greenwashing} and \ref{def:greenwashing2}, greenwashing is not defined by easily identifiable properties but as a general concept. Since the concept is so unspecific, researchers focused on components indicative of potentially misleading communications but easier to define.  
Because of this, in this review we are mentioning ``greenwashing'' explicitly, but also paraphrasing it as ``misleading communications'', ``misrepresentation of the company's environmental impact/stance/performance'', or mentioning only components of it such as ``cheap talk'', ``selective disclosure/transparency'', ``deceptive techniques'', ``biased narrative''. They should all be understood in the context of climate-related misleading communications, as components associated or indicative of potential greenwashing even if they are not synonymous.

\section{Foundational Tasks for Greenwashing Detection}
\label{sec:intermediary tasks}
\subsection{Pretraining Models on Climate-Related Text}
\label{sec:domain specific model}

The first step in applying a language model for a given task is typically the pretraining process, which involves training the model on relevant corpora. Although general-purpose models such as BERT \citep{devlin-etal-2019-bert} and LLaMA \citep{touvron2023llamaopenefficientfoundation} have demonstrated strong performance in various tasks, they are inherently limited by the knowledge and vocabulary present in their training corpora. This has motivated the development of climate-focused language models, which aim to encode domain knowledge directly into the pretrained representations.

\task A language model is pretrained on climate-related text, to produce a domain-specific language model that will be subsequently fine-tuned to specific tasks. There are two approaches: 
\begin{enumerate} 
\item[\textit{(i)}] Training a model with a domain-specific corpus from scratch, or 
\item[\textit{(ii)}] Further-training\footnote{Further-training refers to the process of training a pretrained model on its original task using additional domain-specific data to specialize its knowledge for that domain.} a generalist pretrained model on a domain-specific corpus.
\end{enumerate}

\datasets A variety of climate-related corpora have been assembled for this purpose, ranging from scientific abstracts and policy documents to sustainability reports and ESG disclosures. These corpora differ in scope and quality, and only a few are publicly accessible. The lack of standardized, openly available pretraining datasets remains a barrier to consistent benchmarking and reproducibility. \citet{nicolas_webersinke_climatebert_2021} gathered climate-related news articles, research abstracts, and corporate reports. 
\citet{vaghefi2022deep} introduced Deep Climate Change, a dataset composed of abstracts of articles from climate scientists, and built a corpus specific to climate research texts. 
\citet{schimanski_bridging_2023} introduced a dataset that focuses on text related to Environment Social and Governance (ESG). Similarly, \citet{Mehra_2022} built a dataset using text from the Knowledge Hub of Accounting for Sustainability for an ESG domain-specific corpus.
More recently, \citet{thulke2024climategpt} proposed a dataset comprising news, publications (abstracts and articles), books, patents, the English Wikipedia, policy and finance-related texts, Environmental Protection Agency documents, and ESG, and IPCC reports. \citet{mullappilly-etal-2023-arabic} proposed a climate-specific multilingual dataset. \citet{yu_climatebug_2024} published a pretraining dataset constructed with annual and sustainable reports from EU banks. 
Unfortunately, to the best of our knowledge, only \citep{yu_climatebug_2024}'s ClimateBUG-data dataset is publicly available. \smallskip

\solutions Based on the above datasets, the models ClimateBERT, ClimateGPT-2, climateBUG-LM and EnvironmentalBERT, ESGBERT were proposed \citep{nicolas_webersinke_climatebert_2021, vaghefi2022deep, yu_climatebug_2024, schimanski_bridging_2023, Mehra_2022}. 
More recently, generative domain-specific models such as Llama-2 for ClimateGPT \citep{thulke2024climategpt} or Vicuna for Arabic Mini-ClimateGPT \citep{mullappilly-etal-2023-arabic} have also been proposed. \smallskip

\paragraph{Performance of Models} The domain-specific models drastically improve the performance on domain-specific masked-language modeling \citep{nicolas_webersinke_climatebert_2021, yu_climatebug_2024} and next token prediction \citep{vaghefi2022deep}. They were also evaluated on domain-specific downstream tasks. These tasks are either based on pre-existing datasets such as ClimateFEVER \citep{diggelmann_climate-fever_2020} or introduced by the authors, as in ClimateBERT's climate detection \citep{nicolas_webersinke_climatebert_2021}. At this stage, we evaluate whether fine-tuning enhances downstream performance, with a detailed analysis of the tasks presented in subsequent sections.\smallskip

\begin{table}[ht]
    \centering
    \caption{Performance of ClimateBERT on domain-specific tasks (using F1-score, except for TCFD classification which is using ROC-AUC). The figure displayed in the tables are the values reported by authors in the corresponding studies; with the following abbreviations: \textit{CL} for Climate, \textit{Dis} for Distil, \textit{Evid.} for evidences. Each row reports the performances of models fine-tuned in the same experimental setting.
        Detailed performances are reported in Appendix \ref{app:perf}.
    }
    \label{tab:comparison climatebert}
    \begin{tabular}{lccc}%
        \toprule
        & \textbf{Cl. BERT} & \textbf{Dis. RoBERTa} & \textbf{NB} \\
        \midrule
        Climatext \citep{spokoyny2023answering, varini_climatext_2020} & 85.14 & \textbf{86.06} & 83.39 \\
        Climate Detection \citep{nicolas_webersinke_climatebert_2021} & \textbf{99.1}$\pm1$ & 98.6$\pm0.8$ & \\
        %Climate Detection \citep{bingler2023cheaptalkspecificitysentiment} & $97$ & & $87$ \\
        Sentiment \citep{nicolas_webersinke_climatebert_2021} & \textbf{83.8}$\pm3.6$ & 82.5$\pm4.6$ & \\ 
        % Sentiment \citep{bingler_cheap_2021} & 80 & & 72 (+5) \\
        Net-zero/Reduction \citep{tobias_schimanski_climatebert-netzero_2023} & \textbf{96.2}$\pm0.4$ & 94.4$\pm0.6$ \\ 
        \bottomrule
        \toprule
         & \textbf{Cl. BERT} & \textbf{Dis. RoBERTa} & \textbf{TF-IDF} \\
        \midrule
        TCFD classification \citep{sampson_tcfd-nlp_nodate} & 0.852 & 0.819 & \textbf{0.867} \\
        \bottomrule
        \toprule
         & \textbf{Cl. BERT} & \textbf{Dis. RoBERTa} & \textbf{SVM} \\
        \midrule
        SciDCC \citep{spokoyny2023answering, mishra2021neuralnere} & \textbf{52.97} & 51.13 & 48.02 \\
        ClimaTOPIC \citep{spokoyny2023answering} & \textbf{64.24} & 63.61 & 58.34 \\
        Cl.FEVER (claim) \citep{xiang_dare_2023, diggelmann_climate-fever_2020} & \textbf{76.8} & 72 & \\
        Cl.FEVER (claim) \citep{nicolas_webersinke_climatebert_2021, diggelmann_climate-fever_2020} & \textbf{75.7}$\pm4.4$ & 74.8$\pm3.6$ & \\
        Cl.FEVER (evid.) \citep{spokoyny2023answering, diggelmann_climate-fever_2020} & \textbf{61.54} & \textbf{61.54} & \\
        Cl.Stance \citep{spokoyny2023answering, vaid-etal-2022-towards} & \textbf{52.84} & 52.51 & 42.92 \\
        ClimateEng \citep{spokoyny2023answering, vaid-etal-2022-towards} & 71.83 & \textbf{72.33} & 51.81 \\
        ClimaINS \citep{spokoyny2023answering} & \textbf{84.80} & 84.38 & 86.00 \\
        ClimaBENCH \citep{spokoyny2023answering} & \textbf{69.44} & 69.27 & \\
        Nature \citep{Schimanski2024nature} & \textbf{94.11} & 94.03 & \\
        \bottomrule
        \toprule
        & \textbf{Cl. BERT} & \textbf{SVM+ ELMo} & \textbf{SVM+ BoW} \\
        \midrule
        Commitment\&Actions \citep{bingler2023cheaptalkspecificitysentiment} & \textbf{81} & 79 & 76 \\
        Specificity \citep{bingler2023cheaptalkspecificitysentiment} & \textbf{77} & 76 & 75 \\
        \bottomrule
        \toprule
         & \textbf{Cl. BERT} & \textbf{Dis. BERT} & \textbf{SVM} \\
        \midrule
        Env. Claim \citep{stammbach_environmental_2023} & \textbf{83.8} & 83.7 & 70.9 \\
        \bottomrule
        \toprule
         & \textbf{Cl. BERT} & \textbf{Dis. BERT} & \textbf{LSTM} \\ %\textbf{POS-Bi-LSTM-Attention} \\
        \midrule
        DARE sentiment \citep{xiang_dare_2023} & 87.4 & \textbf{89.9} & 88.2 \\
        \bottomrule
    \end{tabular}%

\end{table}

\begin{table}[ht]
    \centering
    \caption{F1-scores of multiple models compared to EnvironmentalBERT. The figure displayed in the tables are the values reported by authors in the corresponding studies; with the following abbreviations: \textit{Dis} for Distil, \textit{EnvDisRob} for EnvDistilRoBERTa. Each row reports the performances of models fine-tuned in the same experimental setting.
        Detailed performances are reported in Appendix \ref{app:perf}.
    }
    \label{tab:comparison env}
    \begin{tabular}{lcccc}
        \toprule
        & \textbf{RoBERTa} & \textbf{EnvRoBERTa} & \textbf{DisRoBERTa} & \textbf{EnvDisRoB.} \\
        \midrule
        Environment \citep{schimanski_bridging_2023} & 92.35$\pm2.29$ & \textbf{93.19}$\pm1.65$ & 90.97$\pm2$ & $92.35\pm1.65$ \\
        Social \citep{schimanski_bridging_2023} & 89.87$\pm1.35$ & \textbf{91.90}$\pm1.79$ & 90.59$\pm1.03$ & $91.24\pm1.86$ \\
        Governance \citep{schimanski_bridging_2023} & 77.03$\pm1.82$ & 78.48$\pm2.62$ & 76.65$\pm2.39$ & \textbf{78.86}$\pm1.59$ \\
    \bottomrule
        \toprule
         & \textbf{EnvDisRoB.} & \textbf{ClimateBERT} & \textbf{RoBERTa} & \textbf{DisRoBERTa} \\
        \midrule
        Water \citep{Schimanski2024nature} & 94.47$ \pm 1.37$ & \textbf{95.10}$ \pm 1.13$ & 94.55$ \pm 0.86$ & $94.98 \pm 1.16$ \\
        Forest \citep{Schimanski2024nature}   & \textbf{95.37}$ \pm 0.92$ & 95.34$ \pm 0.94$ & 94.78$ \pm 0.48$ & 95.29$ \pm 0.65$ \\
        Biodiversity \citep{Schimanski2024nature}   & \textbf{92.76}$ \pm 1.01$ & 92.49$ \pm 1.03$ & 92.46$ \pm 1.54$ & 92.29$ \pm 1.23$ \\
        Nature \citep{Schimanski2024nature}  & \textbf{94.19}$ \pm 0.81$ & 93.50$ \pm 0.64$ & 93.97$ \pm 0.26$ & 93.55$ \pm 0.72$ \\
        \bottomrule
        \end{tabular}
    
\end{table}

\begin{table}[ht]
    \centering
    \caption{F1-scores of ClimateGPT-2 and custom performances of ClimateGPT compared to similarly sized models on domain-specific benchmarks. The figure displayed in the tables are the values reported by authors in the corresponding studies.
        Detailed performances are reported in Appendix \ref{app:perf}.
    }
    \label{tab:climateGPT}
    \begin{tabular}{lcc}
        \midrule
         & \textbf{GPT-2} & \textbf{climateGPT-2} \\
        \midrule
        ClimateFEVER \citep{vaghefi2022deep} & 62 & \textbf{72} \\
        \midrule
        \midrule
         & \textbf{LLama-2-Chat (7B)} & \textbf{ClimateGPT (7B)} \\
        \midrule
        Multiple Benchmarks \citep{thulke2024climategpt} & 71.4 & \textbf{77.1} \\
        \midrule
    \end{tabular}
\end{table}

As shown in Table~\ref{tab:comparison climatebert} and Table~\ref{tab:comparison env}, evaluations on climate-related downstream tasks show mixed but informative results:  
(i) domain-adapted models often outperform their base counterparts on specialized tasks,
(ii) gains are frequently modest, and  
(iii) strong generalist baselines—when fine-tuned on the same domain-specific datasets—remain highly competitive.  
These findings indicate that domain adaptation helps encode environmental terminology and discourse structure, but its benefits are task-dependent and not uniformly substantial.
In Table~\ref{tab:climateGPT}, both ClimateGPT-2 and ClimateGPT significantly outperform the baseline model. These models benefit from additional training on climate-related data, highlighting the advantages of domain adaptation. However, when trained from scratch, \citet{thulke2024climategpt} report substantially lower performance, underscoring the importance of large-scale pretraining data.

\paragraph{Insights} In theory, domain-specific language models should outperform general models by leveraging their tailored vocabulary and contextual understanding. However, our findings suggest that the improvements provided by these models on the proposed domain-specific downstream tasks are limited. Indeed, we found that fine-tuning general models on domain-specific datasets consistently yields competitive performance. These findings do not undermine the significance of existing works that have been fundamental in demonstrating the potential of domain-specific models, particularly in generating domain-specific text. Moreover, ClimateGPT \citep{thulke2024climategpt}
demonstrated significant improvement for the domain-adapted Llama-2, showing that domain-adaptation is highly relevant. Future research could focus on designing tasks where a lack of domain understanding—whether in terms of vocabulary or knowledge—significantly hinders performance.
\subsection{Climate-Related Topic Detection}
\label{sec:climate-related topic}

The first step toward detecting climate-related greenwashing is identifying text addressing climate-related topics.

\task Given an input sentence or a paragraph, output a binary label,  ``\textit{climate-related}'' or ``\textit{not climate-related}''.

\begin{table}[ht]
\centering
\caption{Label definitions of the datasets related to climate change and sustainability topic detection.}
\label{tab:guidelines climate-related}
\begin{tabular}{p{3cm}p{3cm}p{1.5cm}p{4cm}}
\toprule
\textbf{Dataset} & \textbf{Input} & \textbf{Labels} & \textbf{Positive Label Details}  \\ \midrule
\raggedright ClimateBug-data \citet{yu_climatebug_2024} & \raggedright Sentences from bank reports  & \textit{relevant/} \textit{irrelevant} & Climate change and sustainability (including ESG, SDGs related to the environment, recycling and more) \\ \midrule
\raggedright ClimateBERT's climate detection \citet{bingler2023cheaptalkspecificitysentiment} & \raggedright  Paragraphs from reports & \textit{1/0} & Climate policy, climate change or an environmental topic \\ \midrule
\raggedright Climatext (Wikipedia, 10-K, claims) \citet{varini_climatext_2020} & \raggedright  Sentences from Wikipedia, 10-Ks or web scraping & \textit{1/0} & Directly related to climate-change \\ \midrule
\raggedright Climatext (wiki-doc) \citet{varini_climatext_2020} & \raggedright  Sentences from a Wikipedia page & \textit{1/0} & Extracted from a Wikipedia page related to climate-change \\ \midrule
\raggedright Sustainable signals's reviews \citet{linSUSTAINABLESIGNALSijcai2023} & \raggedright  Online product reviews (user comments) & \textit{relevant/} \textit{irrelevant} & Contains terms related to sustainability  \\ \bottomrule
\end{tabular}
\end{table}

\zeroshot The earliest work on identifying corporate climate-related text was conducted by \citet{doran_risk_disclosure}, where the authors collected a large corpus of 10-K filings from 1995 to 2008, and filtered climate-related reports using hand-selected keywords. While \citet{doran_risk_disclosure} focused on identifying climate-related text within company reports to assess how businesses addressed climate change, other early efforts took a different approach by analyzing the external impacts of media attention toward climate change on companies. For instance, \citet{Engle_hedging_climatechange} developed the WSJ Climate Change News Index based on identifying climate-related news articles in media sources using climate-change vocabulary frequency. This index served as a tool to measure the influence of climate-change attention in the media on companies. \citet{csr_report_greenwashing} proposed measuring the environmental content of CSR (corporate social responsibility) reports using a lexicon-based approach, with the goal of comparing CSR reports of environmental violators to companies with clean records.

\datasets More recent studies have introduced several annotated datasets aimed at climate-related corporate text classification. One of the most prominent is ClimaText \citep{varini_climatext_2020}, a large dataset from diverse sources such as Wikipedia, U.S. Security and Exchange Commission 10-K filings, and web-scraped content. The dataset is designed to cover a broad range of topics and document types to help assess climate-related discussions across different domains. It is divided into multiple subsets: wiki-doc, Wikipedia, 10-k, claims. Climatext (Wiki-doc) is annotated automatically, while Climatext (Wikipedia, 10-k and claims) is annotated by humans.
In contrast, ClimateBERT's authors \citep{nicolas_webersinke_climatebert_2021} provide a smaller, more focused dataset consisting of paragraphs from companies' annual and sustainability reports. This dataset was created as a downstream task for evaluating the ClimateBERT model, focusing on corporate disclosures. The dataset was later extended by \citet{bingler2023cheaptalkspecificitysentiment}, increasing its size and refining the annotation to capture both specificity and sentiment related to corporate climate discourse (see Sections~\ref{sec: climate risk},~\ref{sec: claim characteristics}).
Another large dataset, ClimateBUG-data \citep{yu_climatebug_2024}, also focuses on corporate communications but in EU banks.\smallskip

Beyond corporate disclosures, Sustainable Signals \citep{linSUSTAINABLESIGNALSijcai2023} introduces a dataset of product reviews, each classified based on its relevance to sustainability. In addition to focusing on consumer perspectives through reviews, this dataset also examines sustainability aspects in product descriptions, providing a broader understanding of how sustainability is communicated in the context of consumer products.

\paragraph{Labels and Guidelines} As shown in Table \ref{tab:guidelines climate-related}, although these datasets address similar tasks, they differ significantly in their label definitions. The scope of the labels varies, ranging from a narrow focus on climate change to broader topics such as sustainability and environmental impact. 
While the task of the Climatext (wiki-doc) dataset is to detect whether a sentence is related to climate change or not, the weak label actually means that the sentence originates from a Wikipedia page related to climate change. Which is not the same task, as it focuses on the source of the statement rather than the actual content or its relevance to climate-related topics. However, the weakly labeled dataset serves as a useful filtering mechanism, helping to identify sentences that are potentially interesting. It provides an initial pool of data that can be refined with human annotations. Human annotations ensure that the labels align with the task’s goals, something weak labels alone cannot achieve.

\solutions The solutions proposed to tackle that task range from keyword-based models \citep{varini_climatext_2020, bingler2023cheaptalkspecificitysentiment} to fine-tuning BERT-like models \citep{varini_climatext_2020, nicolas_webersinke_climatebert_2021, garridomerchán2023finetuning, bingler2023cheaptalkspecificitysentiment, yu_climatebug_2024}. The performance of the best models is above 90\% \citep{garridomerchán2023finetuning, bingler2023cheaptalkspecificitysentiment, yu_climatebug_2024}.\footnote{See Table~\ref{tab:reported perf climate} in the appendix for details.}

\paragraph{Insights}
The task of determining whether a statement is climate-related has been extensively studied, with numerous datasets available to support research in this area. State-of-the-art fine-tuned models now achieve near-perfect performance on these datasets, indicating that, under current conditions, the task is effectively solved.
However, the labels used across the studies vary in scope. 
This highlights the inherent challenge of isolating climate-related discourse, given its close association with broader environmental and sustainability topics.
For example, whether to include only explicit references to climate change or incorporate broader environmental and sustainability discussions, can profoundly influence the results and interpretations of such analyses.
Moreover, relying only on weak labels might result in tasks that 
fail to capture the complexities of the subject.

\subsection{Thematic Analysis}
\label{sec:sub-topics}

Once a text is known to be related to climate, one aims to know the exact topic of the text.
Companies may prioritize specific categories while underreporting others, potentially signaling selective transparency practices such as greenwashing or greenhushing.

\task Given an input sentence or paragraph, output a subtopic related to climate change. This is a multiclass classification task that can be supervised or unsupervised, depending on the availability of labeled data. Alternatively, it can be framed as a clustering task with the goal of discovering latent subtopic structures.

    \begin{table}[ht!]
\centering
\caption{Summary of datasets, labels, scopes, and sources for climate-related text classification tasks.}
\label{tab:datasets_subtopic}
\begin{tabular}{p{2.5cm}p{2.5cm}p{3cm}p{4cm}}
\toprule
\textbf{Dataset} & \textbf{Input} & \textbf{Labels}                                                                                  & \textbf{Label details}                                          \\
\midrule
TCFD rec. \citet{bingler_cheap_2021} & \raggedright  Paragraphs from corporate annual reports & \multicolumn{2}{p{7cm}}{TCFD 4 main categories: \textit{Metrics and Targets, Risk Management, Strategy, Governance and General} (see appendix \ref{app:tcfd} for details)}             \\
TCFD \citet{sampson_tcfd-nlp_nodate} & \raggedright Paragraphs from  regulatory and discretionary reports & \multicolumn{2}{p{7cm}}{TCFD 11 recommendations: \textit{Metrics and Targets (board’s oversight,  management’s role), Risk Management (identified risk, impact, resilience), ...} (see Appendix \ref{app:tcfd} for details)}             \\ 
\midrule
FineBERT's ESG \citet{huangFinBERTLargeLanguage2020} & Sentences from 10-K  & \textit{Environmental, Social, Governance, General} & Environmental - e.g., climate
change, natural capital, pollution and waste, and environmental opportunities \\
ESGBERT's ESG \citet{schimanski_bridging_2023} & \raggedright Sentences from reports and corporate news & \textit{Environment, Social, Governance and None} & Environmental criteria comprise a
company’s energy use, waste management, pollution, [...]
as well as compliance with governmental regulations.
Special areas of interest are climate
change and environmental sustainability.  \\
Multilingual ESG classification \citet{LEE2023119726} & \raggedright Sentences from Korean reports (English and Korean)  & \textit{Environment, Social,
Governance, Neutral, and Irrelevant} & Environmental factors include the reduction of hazardous substances, eco-friendly management, climate change, carbon emission, natural resources, [...] \\ 
\midrule
SciDCC \citet{mishra2021neuralnere} & \raggedright News articles (Title, Summary, Body)  & \textit{Environment, Geology, Animals, Ozone Layer, Climate, etc.} & Category in which the article was published (Automatic Label)   \\
Transaction Ledger \citet{jain_supply_2023} & Transaction ledger entry (description of transaction) & \raggedright \textit{Accounting Adjustments, Administration, Advertising, Benefits \& Insurance, etc.} & Standardized commodity classes  \\
ClimateEng \citet{vaid-etal-2022-towards}  & Tweets posted during COP25 filtered by keywords (relevant to climate-change)  & \raggedright \textit{Ocean/Water, Politics, Disaster, Agriculture/Forestry, General} & Sub-categories of climate-change\\
ESGBERT's Nature \citet{Schimanski2024nature} & \raggedright Paragraphs from reports  & \raggedright \textit{General, Nature, Biodiversity, Forest, Water} & Multi-label Nature-related topics                   \\
ClimaTOPIC \citet{spokoyny2023answering} & \raggedright CDP responses (short texts) & \raggedright \textit{Adaptation, Buildings, Climate Hazards, Emissions, Water, etc.} & Category of the question (Automatic Label)          \\
\bottomrule
\end{tabular}

\end{table}

\subsubsection{Classifying Climate-Related Financial Disclosure (TCFD)}
\label{sec: tcfd}

The Task Force on Climate-Related Financial Disclosures (TCFD) proposes four main categories of climate-related disclosure (Governance, Strategy, Risk Management, and Metrics and Targets) and 11 recommendations for climate-related disclosures.\footnote{see Appendix \ref{app:tcfd} for details.} Finding texts associated to these categories in company communication can help identify
gaps and inconsistencies in reporting.

\zeroshot \citet{dingCarbonEmissionsTCFD2023} introduced a score to quantify the extent of climate-related content in corporate disclosures, as well as 4 TCFD-category-based similarity scores. The score is based on sentence similarity with sentences from TCFD referential documents. They identified discrepancies that could signal greenwashing by comparing these scores with actual carbon emission data. \citet{auzepy_evaluating_2023} examined reports from banks endorsing the TCFD recommendations. They proposed a fine-grained analysis of TCFD categories using a zero-shot entailment classification method on sentences.

\datasets Although unsupervised methods can be employed for thematic analysis, recent studies have demonstrated the value of curated datasets for categorizing corporate disclosures. For instance, \citet{bingler_cheap_2021} leveraged report headings to annotate paragraphs into the four main TCFD categories. Similarly, \citet{sampson_tcfd-nlp_nodate} assessed the quality of climate-related disclosures using a dataset of 162k sentences, where each sentence was labeled according to the TCFD recommendations by experts through a question answering (QA) process.

\subsubsection{Classifying Environmental, Social, and Governance Disclosure (ESG)}

Beyond climate-specific reporting, Environmental, Social, and Governance (ESG) disclosures represent a broader set of non-financial metrics that evaluate a company's sustainability performance. ESG reporting has become a global standard, encompassing climate-related initiatives and other dimensions of corporate responsibility. The standardized nature of ESG metrics makes them valuable for computational models that aim to detect inconsistencies in corporate claims, such as overstating environmental impacts through greenwashing. However, as noted in \citet{berg2022aggregate}'s aggregated analysis of ESG ratings, discrepancies persist across different evaluation frameworks. These inconsistencies raise the question of whether artificial intelligence can be harnessed to improve the consistency and reliability of ESG assessments.

\zeroshot \citet{rouenEvolutionESGReports2023} proposed an ML algorithm to describe the content of ESG reports. They used industry-topic dictionaries to compute the topic frequency using a TF-IDF-based algorithm. They used that classification to identify selective disclosure. \citet{Mehra_2022} experimented with sentence similarity to extract sentences that were the most relevant to environmental factors, and \citet{bronzini_glitter_2023} selected sentences related to ESG using INSTRUCTOR-xl \citep{su2023embeddertaskinstructionfinetunedtext}. They both subsequently used the extracted sentences to predict ESG scores.

\datasets Multiple works \citet{huangFinBERTLargeLanguage2020, schimanski_bridging_2023, LEE2023119726} proposed datasets for topic classification with the labels: Environment, Social, Governance and General.
\citet{huangFinBERTLargeLanguage2020} focused on reports while \citet{schimanski_bridging_2023} added corporate news. \citet{schimanski_bridging_2023} introduced a larger dataset for pre-training domain-specific models and an evaluation dataset of 2,000 sentences. While most ESG-related reports are typically available in English, some local companies provide them only in their native language. To partially address this, \citet{LEE2023119726} proposed a multilingual dataset constructed from Korean corporate reports.

\subsubsection{Other Topics}
\label{sec:other-sub-topics}

Although TCFD and ESG are established frameworks for analyzing corporate communications, many other topics may be of interest. These topics can follow other frameworks such as CDP, but the thematic analysis can also be unsupervised, effectively discovering the topics mentioned.

\zeroshot \citet{hyewon_kang_analyzing_2022} analyzed the content of sustainable reports with thematic analysis. It was conducted using sentence similarity between the reports and content from the SDGs website. \citet{yu_climatebug_2024} examined the themes covered in the climate-related sections of bank reports by employing a dual approach: using hand-selected keywords to target anticipated topics and applying clustering techniques to group climate-related sentences and uncover underlying themes.  \citet{bjarne_brie_mandatory_2022} leveraged ClimateBERT to extract climate-related paragraphs from corporate documents. They then applied a Structural Topic Model (STM) \citep{STM} to identify thematic clusters within the data. This STM-based approach was also used by \citet{fortes2020tracking} to identify influential topics for EUR/USD exchange rate.
Their findings highlighted a significant discrepancy between the European Central Bank (ECB) and the Federal Reserve (FED), with the ECB attributing greater importance to climate change in the context of financial stability than the FED. Such thematic analysis can be used to identify selective disclosure.

\datasets \citet{spokoyny2023answering} introduced ClimaTOPIC, a topic classification dataset derived from responses to the CDP questionnaires, organized into 12 categories that align with the thematic structure of the CDP questions. This can be used to identify sections relevant to a particular CDP topic (e.g., Emissions, Adaptation, or Energy), enabling the extraction of structured information from unstructured documents. 
\citet{Schimanski2024nature} presents a dataset focused on nature-related topics (such as Water, Forest, and Biodiversity) extracted from annual reports, sustainability reports, and earnings call transcripts (ESGBERT Nature). While less explicitly linked to climate change, these topics often reflect the consequences of environmental degradation associated with climate issues. 
While NLP is often used to analyze corporate report texts, it can also be used to analyze tables. For example, \citet{jain_supply_2023} proposes estimating Scope 3 emissions \footnote{Scope 3 emissions refer to all indirect greenhouse gas (GHG) emissions that occur in a company’s value chain, excluding emissions from the company’s own operations (Scope 1) and its purchased electricity (Scope 2).} based on corporate transactions (see Section~\ref{sec:env prediction} for details). The first step involves grouping expenses into standardized commodity classes, for which they introduced a dataset of 8K examples of corporate expenses categorized accordingly. 
As previously discussed, social media and news articles offer a broader perspective for cross-referencing corporate communications. For example, \citet{vaid-etal-2022-towards} introduced ClimateEng, a climate-related topic classification dataset composed of tweets grouped into five categories: Disaster, Ocean/Water, Agriculture/Forestry, Politics, and General. 
This dataset can be used to detect greenwashing by comparing the public’s concerns with the narrative presented in corporate sustainability reports. Similarly, \citet{mishra2021neuralnere} published the Science Daily Climate Change dataset (SciDCC), consisting of 11,000 news articles grouped into 20 climate-related categories. This dataset offers another avenue for identifying greenwashing, as companies’ communications can be cross-referenced with external news coverage, potentially highlighting contradictions or omissions in corporate reports.

\subsubsection{Models and Conclusion}

\solutions The most common solution proposed for these tasks is fine-tuning a Transformer model \citep{huangFinBERTLargeLanguage2020, schimanski_bridging_2023, LEE2023119726, bingler_cheap_2021, sampson_tcfd-nlp_nodate, vaid-etal-2022-towards, spokoyny2023answering, Schimanski2024nature, jain_supply_2023}. To contextualize the performance of Transformer models, some studies also report performances of classical baselines \citep{huangFinBERTLargeLanguage2020, spokoyny2023answering, Schimanski2024nature, jain_supply_2023, bingler_cheap_2021, sampson_tcfd-nlp_nodate}. \citet{bingler_cheap_2021} proposed a custom approach combining logistic regression with features from a fine-tuned language model, while \citet{sampson_tcfd-nlp_nodate} also evaluated clustering techniques and stacked models. The performance of the best-performing model is systematically high (above 80\%), except on SciDCC and ClimaTOPIC.
Classical approaches (SVM, NB) also reach good performances, under-performing by less than 20\% on all datasets (except on \citet{bingler_cheap_2021}'s dataset). The TF-IDF baseline even outperformed fine-tuned Transformers on \citet{sampson_tcfd-nlp_nodate}'s dataset. This shows that those topics have distinguishable vocabularies. 

\paragraph{Automatic Labels} The performance gap observed in \citet{bingler_cheap_2021} may be attributed to using paragraph-level inputs rather than sentence-level inputs aggregated for classification, as fine-tuned models also achieve around 20\% precision at the paragraph level.
The lower performances on SciDCC\citet{mishra2021neuralnere} and ClimaTOPIC\citet{spokoyny2023answering} are likely caused by the automation of the labeling process. The labels are therefore not designed to be labels. In SciDCC there are labels that are highly similar (e.g. \textit{Endangered Animals} and \textit{Extinctions}), or labels that include other labels (e.g. \textit{Environment}, \textit{Climate}, \textit{Pollution}). As they are categories from a journal, the categories changed in time. For ClimaTOPIC, the labels are question categories, which are designed to group questions not to precisely identify them. Therefore, a question about "the emission of a building" might fit in either \textit{Emissions} or \textit{Building}, yet it is assigned only one label. 

\paragraph{Insights} Topic classification is a well-established area of research in natural language processing (NLP) and has been extensively applied to climate-related topics. Fine-tuned Transformer models proposed in the literature consistently demonstrate near-perfect performance on human-annotated datasets, indicating that the task is largely solved. When reported, classical keyword baselines also perform relatively well, achieving significantly better than random performance and approximately 80\%  of the performance of fine-tuned models, suggesting the topics have distinct vocabularies. Future research could investigate topics that are lexically similar but distinct in their subject or focus.
Topic detection serves as a valuable tool for structuring documents and has been utilized to analyze the extent to which companies address specific topics, enabling the detection of selective disclosure \citet{bingler2023cheaptalkspecificitysentiment, bingler_cheap_2021}. However, it is critical to ensure that labels are well-defined and data is accurately annotated. 
Lower performance observed on automatically generated labels may reflect challenges in predictability, potentially due to difficulties caused by the automatic annotation. These works might be revisited to conduct human annotations. 

\subsection{In-depth Disclosure: Climate Risk Classification}
\label{sec: climate risk}

Climate change can bring both risks and opportunities for companies. Potential risks are, e.g., reputational risks (e.g. environmental controversies), regulatory risks (e.g. new regulations on emissions), and physical risks (e.g. droughts impacting production).  Opportunities are, e.g., financial opportunities (e.g. benefiting from grants that aim to support less polluting industries), market opportunities (e.g. electric cars becoming more popular with environmentally conscious clients), etc. If a company systematically avoids discussing climate-related risks or disproportionately emphasizes opportunities, it creates a biased narrative that can serve as an indicator of greenwashing.

\task Given an input sentence or a paragraph, output ``opportunity'' or ``risk'' label. Some works focus only on risks, classifying them into types of risks (e.g., physical risk, reputational risk, regulatory risk, or transition risk)

\begin{table}[ht!]
\centering
\caption{Summary of datasets, labels, scopes, and sources for climate-related Risk classification.}
\label{tab:datasets_risk}
\centering
\begin{tabular}{p{2cm}p{3cm}p{3cm}p{3cm}}
\toprule
\textbf{Dataset} & \textbf{Input}   &  \textbf{Labels}                                                                                  &                                       \\
\midrule
Ask BERT's Climate Risk \citet{kolbel_ask_2021} & \raggedright Sentences from TCFD's example reports and non-climate-related sentences &  \multicolumn{2}{p{7cm}}{Risk type: \textit{Transition risk, physical risk, and general risk (no guidelines)}}             \\ \midrule
Climate Risk \citet{Friederich_climate_risk_disclosure} & Paragraphs from European companies annual reports & \multicolumn{2}{p{7cm}}{Risk type:\textit{Acute, Chronic, Policy \& legal, Tech \& Market, Reputational, and Negative} (no guidelines)}       \\ \midrule
ClimateBERT's Sentiment \citet{bingler2023cheaptalkspecificitysentiment}  & Paragraphs from companies' annual reports            & \multicolumn{2}{p{7cm}}{\textit{Risk} or threat that negatively impacts an entity of interest (negative sentiment); or \textit{Opportunity} arising due to climate change (positive sentiment); \textit{Neutral} otherwise.} \\ \midrule
Sentiment Analysis \citet{xiang_dare_2023} & Paragraphs from academic texts on climate change and health published between 2013 and 2020  & \multicolumn{2}{p{7cm}}{\textit{Risk} (negative) if it discusses climate change causing public health issues, serious consequences, or worsening trends, greenwashing. \textit{Opportunity} (positive): highlights potential benefits, positive actions, or research addressing gaps. \textit{Neutral} otherwise.} \\
\bottomrule
\end{tabular}
\end{table}

\zeroshot Similarly to climate-related detection, climate risk classification has been tackled with keyword-based approaches. \citet{liCorporateClimateRisk2020}, \citet{kheradmand2021a} and \citet{chou_ESG} proposed using dictionaries of words related to climate risk to identify paragraphs dealing with climate risk. \citet{liCorporateClimateRisk2020} constructed risk measures based on the frequency of the terms in the risk dictionaries. \citet{chou_ESG} analyzed the topics mentioned in conjunction with physical and transition risks. \citet{SAUTNER_cliamte_change_exp} proposed using one dictionary for risk and one for opportunity classification.

\datasets Several studies have introduced specialized annotated datasets for risk/opportunity classification. For instance, \citet{kolbel_ask_2021} created a dataset for climate-related risk classification with three categories: physical risk, transition risk, and general risk, using active learning for annotation. Similarly, \citet{Friederich_climate_risk_disclosure} developed an annotated dataset for risk classification with five labels, covering acute physical risk, chronic physical risk, policy and legal risks, technology and market risks, and reputational transition risks. \citet{bingler2023cheaptalkspecificitysentiment} released a dataset as part of the ClimateBERT downstream tasks, focusing on classifying paragraphs from corporate reports into three categories: opportunity, neutral, or risk. Extending beyond corporate disclosures, \citet{xiang_dare_2023} compiled a climate-related risk/opportunity classification dataset of academic texts from the Web of Science and Scopus.

\solutions The solutions proposed are similar to the other tasks: fine-tuned Transformer models \citet{hyewon_kang_analyzing_2022, kolbel_ask_2021, Friederich_climate_risk_disclosure, bingler2023cheaptalkspecificitysentiment, nicolas_webersinke_climatebert_2021, xiang_dare_2023} and keyword-based-features with simple models \citet{kolbel_ask_2021, Friederich_climate_risk_disclosure, bingler2023cheaptalkspecificitysentiment}. The notable exception is \citet{xiang_dare_2023}, who evaluated LSTM-based solutions alongside the Transformers models. Most studies \citep{hyewon_kang_analyzing_2022, kolbel_ask_2021, bingler2023cheaptalkspecificitysentiment, nicolas_webersinke_climatebert_2021, xiang_dare_2023} reported high performances, above 80\%. \citet{kolbel_ask_2021} and \citet{bingler2023cheaptalkspecificitysentiment} also reported good performances for the keyword-based baselines (72\% and 84\%). The performance reported by \citet{Friederich_climate_risk_disclosure} on climate risk tasks highlights distinct challenges across subtasks. When tasked with identifying the specific type of risk in sentences already known to be about climate risks, word-based models outperformed fine-tuned Transformers. This suggests that each risk type had a distinct vocabulary. In contrast, when working with a dataset that reflects real-world proportions of risk and non-risk examples—where the data is heavily imbalanced—classical models struggled due to the overlap in vocabulary between general and climate risk-related text. 

\paragraph{Difference between sentiment and risk} It is important to distinguish risk/opportunity classification from sentiment analysis. While risk/opportunity classification is inspired by sentiment analysis \citet{bingler2023cheaptalkspecificitysentiment}, they play a complementary role. A statement such as \textit{``With proactive climate risk management, we are ready to tackle extreme weather disruptions, ensuring resilience''} is classified as positive using a sentiment analysis model \citet{perez2021pysentimiento}, and as mentioning a risk using a risk/opportunity classification model \citet{bingler2023cheaptalkspecificitysentiment}. Sentiment analysis focuses on the form of the statement, while risk is about the content. As the literature on sentiment analysis \citet{Wankhade2022} is broad, the analysis of the tone of the statement can be done using existing models \citet{marco_polignano_nlp_2022, hyewon_kang_analyzing_2022} complementing approaches on risk classification. 

\paragraph{Insights} The performance of fine-tuned models might indicate that the task of identifying texts talking about risks is solved. However, the work by \citet{Friederich_climate_risk_disclosure} shows limitations when experimenting with heavily imbalanced datasets, which actually correspond to real-world settings. Future research could focus on such skewed settings.

\subsection{Green Claim Detection}
\label{sec:green claim}

Identifying greenwashing involves more than just finding text related to climate change, as a company might simply state factual information about the climate without making any commitments on it. Hence, we now focus on detecting \textit{green claims}, i.e., claims that a product, service, or corporate practice either contributes positively to environmental sustainability or is less harmful to the environment than alternatives \citet{stammbach_environmental_2023,vinicius_woloszyn_towards_2021}. The European Commission calls these claims ``environmental claims'':

\begin{table}[ht]
\centering
\caption{Label definitions of the datasets on green claims detection.}
\label{tab:guidelines green claims}
\begin{tabular}{p{2.5cm}p{2.5cm}p{7cm}}
\toprule
\textbf{Dataset}  & \textbf{Input}  & \textbf{Positive Label description}\\ \midrule
\raggedright Green Claims \citet{vinicius_woloszyn_towards_2021}  & Marketing Tweets  & Environmental (or green) advertisements refer to all appeals that include ecological, environmental sustainability, or nature-friendly messages that target the needs and desires of environmentally concerned stakeholders. \\ \midrule
\raggedright Environmental Claims \citet{stammbach_environmental_2023} & \raggedright Paragraph from reports  & Environmental claims refer to the practice of suggesting or otherwise creating the impression [...] that a product or a service is environmentally friendly (i.e., it has a positive impact on the environment) or is less damaging to the environment than competing goods or services [...]
In our case, claims relate to products, services, or specific corporate environmental performance.   \\ \bottomrule
\end{tabular}

\end{table}

\task  Given an input sentence or a paragraph, output a binary label, ``\textit{green claim}'' or ``\textit{not green claim}''.

\datasets The most notable approaches aiming at identifying climate-related claims introduced annotated datasets \citet{vinicius_woloszyn_towards_2021, stammbach_environmental_2023}.
\citet{vinicius_woloszyn_towards_2021} proposed to focus on social media marketing through the detection of green claims in tweets. \citet{stammbach_environmental_2023} proposed a dataset for environmental claim detection in paragraphs from corporate reports.

\solutions Both \citet{stammbach_environmental_2023} and \citet{vinicius_woloszyn_towards_2021} evaluated fine-tuned Transformer models on the datasets. \citet{stammbach_environmental_2023} also evaluated classical approaches and experimented with a model fine-tuned on general claim detection \citet{Arslan_Hassan_Li_Tremayne_2020} applied to their domain-specific task.
\citet{vinicius_woloszyn_towards_2021} and \citet{stammbach_environmental_2023} demonstrated good performances with fine-tuned Transformer (above 84\%), yet they have very different Inter-Annotator Agreement (IAA), with Krippendorff's $\alpha=0.8223$ for \citet{vinicius_woloszyn_towards_2021} and $\alpha=0.47$ for \citet{stammbach_environmental_2023}. This might be explained by the difference in context: marketing tweets might be easier to understand for humans and more self-sufficient compared to paragraphs extracted from reports.
Additionally, \citet{vinicius_woloszyn_towards_2021} experimented with adversarial attack, showing the sensibility to character-swap and word-swap. This shows that the models tend to rely heavily on particular words, showing only a superficial understanding. While it is difficult to generalize that conclusion given the small size of the dataset, it still indicates that BERT-like architecture can over-fit or rely on superficial cues instead of building an accurate representation of the paragraph. 
It is essential to assess model performance in challenging scenarios, where the presence of nonsensical or noisy inputs can reveal the fragility of model comprehension.

\paragraph{Insights}  
Green claims have been analyzed in company reporting \citep{stammbach_environmental_2023} and social media communication \citep{vinicius_woloszyn_towards_2021}. Fine-tuned models can solve the task rather well, even if challenges remain:
\begin{itemize}
\item \textit{Inter annotator agreement:} The annotator agreement remains low for green claim detection in reports \citep{stammbach_environmental_2023}.
    \item \textit{Integration with existing claim detection literature:} While claim detection is a well-established field with extensive literature, studies on environmental claims do not fully connect with existing research on claim detection. In particular, they do not distinguish between Claims, Verifiable Claims (claims that can be checked), and Check-worthy Claims (claims that are interesting to verify) \citep{panchendrarajan2024claim}.
    \item \textit{Evaluation Sensitivity and Real-World Robustness:} The literature shows that fine-tuned models are sensitive to adversarial attacks, meaning small perturbations of the text influence greatly the performance of the classifier. As for all tasks, it is also important evaluate models robustness in real-world settings. Poor data quality might induce perturbations in the texts, reducing the performance of models.
\end{itemize}

\subsection{Green Claim Characteristics}
\label{sec: claim characteristics}

Once we have established that a sentence is climate-related (Section~\ref{sec:climate-related topic}) and that it is a claim about the company (Section~\ref{sec:green claim}), we can endeavor to further classify the claim into fine-grained categories. Table~\ref{tab:guidelines characteristics} shows various characteristics of claims that have been studied.

\begin{table}[ht]
\centering
\caption{Label definitions the datasets related to characterization of green claims.}
\label{tab:guidelines characteristics}
\begin{tabular}{p{2.5cm}p{2.5cm}p{7cm}}
\toprule
\textbf{Dataset}  & \textbf{Input}  & \textbf{Labels}\\ \midrule
Implicit/Explicit Green Claims \citet{vinicius_woloszyn_towards_2021}  & Marketing Tweets  & \textit{Implicit green claims} raise the same ecological and environmental concerns as \textit{explicit green claims} (see definition in Section \ref{sec:green claim}), but without showing any commitment from the company. If the tweet does not contain a green claim then \textit{No Claim}. \\ \midrule
Specificity \citet{bingler2023cheaptalkspecificitysentiment}  & \raggedright Paragraph from reports  & A paragraph is \textit{Specific} if it includes clear, tangible, and firm-specific details about events, goals, actions, or explanations that directly impact or clarify the firm's operations, strategy, or objectives. \textit{Non-specific} otherwise. \\ \midrule
Commitments \& Actions\citet{bingler2023cheaptalkspecificitysentiment} &  \raggedright  Paragraph from reports  & A paragraph is a commitment or an action if it contains targets for the future or actions already taken in the past. \\ \midrule
Net Zero/Reduction \citet{tobias_schimanski_climatebert-netzero_2023} & \raggedright  Paragraph from Net Zero Tracker \citep{netzerotracker2023} & The paragraph 
 contains either a \textit{Net-Zero} target, a \textit{Reduction} target or no target (\textit{None})  \\ \bottomrule
\end{tabular}

\end{table}

\task Given an input sentence or a paragraph labeled as a green claim, output a more fine-grained characterization of the claim. This is a multi-label classification task; the labels can be about the form (e.g. specificity) or the substance (e.g. action, targets, facts). 

\datasets Existing works characterized both the content and the form of claims. On the content dimension, there are datasets for identifying if the claim is about an action or a commitment \citep{bingler2023cheaptalkspecificitysentiment, tobias_schimanski_climatebert-netzero_2023}. \citet{bingler2023cheaptalkspecificitysentiment} introduce a dataset for the identification of commitments and actions (Yes/No), and \citet{tobias_schimanski_climatebert-netzero_2023} released one for the identification of reduction targets (net zero, reduction, general). On the form dimension, \citet{vinicius_woloszyn_towards_2021} characterized claims as implicit or explicit, and \citet{bingler2023cheaptalkspecificitysentiment} annotated their dataset on the specificity of claims (Specific/Not specific). \citet{bingler2023cheaptalkspecificitysentiment} ultimately used the specificity and commitment/action characteristics to identify cheap talks related to climate disclosure.  

\solutions \citet{vinicius_woloszyn_towards_2021, tobias_schimanski_climatebert-netzero_2023} and \citet{bingler2023cheaptalkspecificitysentiment} experimented with fine-tuned Transformers. \citet{bingler2023cheaptalkspecificitysentiment} also evaluated classical baselines (SVM, NB), and \citet{tobias_schimanski_climatebert-netzero_2023} experimented with GPT-3.5-turbo. All approaches reached good performances, in particular fine-tuned Transformers reaching performances above 80\%. The exception remains the Specificity classification with a F1-score of 77\% (close to the baselines). This low performance might be intrinsic to the task. Indeed, humans have disagreement when distinguishing specific and unspecific claims: \citet{bingler2023cheaptalkspecificitysentiment} measured a low IAA on Specificity (Krippendorff's $\alpha=0.1703$). This might indicate that the task is not well defined in the first place. 

\paragraph{Insights} 
The high performances of fine-tuned models indicate that this type of task is solved. Furthermore, the demonstrated ability of GPT-3.5 as a zero-shot classifier \citet{tobias_schimanski_climatebert-netzero_2023} highlights the potential of large language models to classify these characteristics without requiring extensive annotated datasets. 
These characteristics are particularly useful for identifying statements that may indicate greenwashing. For example, \citet{bingler2023cheaptalkspecificitysentiment} utilized the proportion of non-specific commitments as a \textit{Cheap Talk Index} demonstrating their practical applications.
However, characteristics such as specificity are inherently ambiguous and subjective, as evidenced by the low inter-annotator agreement (IAA). Therefore, future research could focus on disambiguating what constitutes a specific statement from a non-specific one more objectively.
\subsection{Green Stance Detection}
\label{sec:stance detection}

Beyond making claims about the environmental impact of their products and processes, companies contribute to environmental discussions and can have an impact on regulatory frameworks through their communications and industry presence. It is thus helpful to understand the stance of an organization on the existence and gravity of climate change, on climate mitigation and adaptation efforts, and on climate-related regulations. 

\begin{table}[ht!]
\centering
\caption{Summary of datasets, labels, scopes, and sources for climate-related stance detection.}
\label{tab:datasets_stance}
\begin{tabular}{p{2.5cm}p{3cm}p{3cm}p{3cm}}
\toprule
\textbf{Dataset} & \textbf{Input} & \textbf{Labels}                                                                                  & \textbf{Label details}                                         \\
\midrule
ClimateFEVER (evidence)  \citep{diggelmann_climate-fever_2020} 
& A claim and an evidence sentence from Wikipedia  & \textit{Support, Refutes, Not Enough Information} & Determines the relation between a claim and a single evidence sentence \\
\midrule
LobbyMap (Stance) \citep{morio2023an} & Page from a company communications (report, press release, ...) & \textit{Strongly supporting, Supporting, No or mixed position, Not supporting, Opposing} & Given the policy and the page, classifies the stance \\
\midrule
Global Warming Stance Detection (GWSD) \citep{luo_detecting_2020}  & Sentences from news about global warming & \multicolumn{2}{p{6cm}}{Stance of the evidence (\textit{Agree, Disagree, Neutral}) toward the claim: Climate-Change is a serious concern.} \\
\midrule
ClimateStance \citep{vaid-etal-2022-towards} & Tweets posted during COP25 filtered by keywords (relevant to climate-change) & \multicolumn{2}{p{6cm}}{Stance towards climate change prevention: \textit{Favor, Against, Ambiguous}. (Stance used as a broad notion including sentiment, evaluation, appraisal, ...)}  \\
\midrule
Stance on Remediation Effort \citep{lai_using_2023} & Texts extracted from the TCFD sections of the financial reports  & \multicolumn{2}{p{6cm}}{The text indicates \textit{support} for climate change remediation efforts or \textit{refutation} of such efforts. (No guidelines)} \\
\bottomrule
\multicolumn{4}{c}{\textbf{Related subtask}} \\
\toprule
ClimateFEVER (claim)  \citep{diggelmann_climate-fever_2020}  & A claim and multiple evidence sentences from Wikipedia & \textit{Support, Refutes, Debated, Not Enough Information} & Determines if a claim is supported by a set of retrieved evidence sentences  \\
\midrule
LobbyMap (Page) \citep{morio2023an} & \multirow{2}{3cm}[-0cm]{Page from a company communication (report, press release, etc)} & \textit{Binary} & Contains a stance on a remediation policy  \\
LobbyMap (Query) \citep{morio2023an} & & \textit{GHG emission regulation, Renewable energy, Carbon tax, etc)} & Classifies the remediation policy \\
\bottomrule
\end{tabular}

\end{table}

\task Given two input sentences or paragraphs, one labeled as the claim and one as the evidence, predict the stance between the two: supports, refutes or neutral. Some studies fix the claim and only vary the evidence (e.g. the claim is always \textit{Climate change poses a severe threat}), training a model to predict the stance of the evidence in respect to the fixed claim. 
Other studies train a model to predict the relation between varying claims/evidences.

\paragraph{Related subtask} The first subtask is collecting the evidences when the claims are already available (e.g. if they were collected manually). In order to build these datasets, research can rely on simple heuristics such as downloading all tweets published during the COP25 filtered with keywords \citet{vaid-etal-2022-towards}. However, other researchers performed a more elaborate procedure. \citet{diggelmann_climate-fever_2020} proposed a pipeline for collecting evidence. To retrieve relevant evidence from Wikipedia for a given claim, the pipeline involves three steps: document-level retrieval using entity-linking and BM25 to identify top articles, sentence-level retrieval using sentence embeddings trained on the FEVER dataset to extract relevant sentences, and sentence re-ranking using a pretrained ALBERT model to classify and rank evidence based on relevance. 
The second one encompasses the broader process: identifying claims and finding evidence, before predicting the relation. \citet{Wang2021EvidenceBA} used a generic claim detection model trained on ClaimBuster \citet{Arslan_Hassan_Li_Tremayne_2020} for the claims and Google Search API to collect evidence. 
Finally, the last subtask is to train multiple models on each step: identifying evidences, identifying the claim and classifying the stance. This is the approach used by \citet{morio2023an}.

\datasets As described previously, researchers built datasets of evidence related to one claim while focusing stance in tweets on seriousness of climate change such as Global Warming Stance Detection (GWSD) \citet{luo_detecting_2020}, ClimateStance \citet{vaid-etal-2022-towards} stance on climate change of tweets posted during the COP25, or \citet{lai_using_2023}'s dataset which focuses on the stance toward climate change remediation efforts.
Another research initiative built a dataset of claim-evidence pairs, ClimateFEVER, and then trained a model on the stance classification \citet{diggelmann_climate-fever_2020, Wang2021EvidenceBA}.
Finally,~\citet{morio2023an} proposed a dataset to assess corporate policy engagement built upon LobbyMap, which tackles the 3 steps: finding pages with evidence, identifying the claims targeted by that evidence, and classifying the stance.

\solutions The solution proposed for to tackle stance classification are fine-tuned Transformer models \citet{vaid-etal-2022-towards, vaghefi2022deep, xiang_dare_2023, morio2023an, nicolas_webersinke_climatebert_2021, Wang2021EvidenceBA, spokoyny2023answering, lai_using_2023, luo_detecting_2020} and classical approaches \citet{spokoyny2023answering, morio2023an, luo_detecting_2020}. \citet{vaid-etal-2022-towards} also experimented with FastText and \citet{xiang_dare_2023} with LSTM-based models 
. The performances are quite heterogeneous, \citet{lai_using_2023} reaching F1-scores around 90\% on classification of stance on remediation efforts, \citet{luo_detecting_2020} around 72\% on GWSD, while performance on ClimateStance could not exceed 60\%. Performances on ClimateFEVER are also quite low, however, when selecting only non-ambiguous examples, \citet{xiang_dare_2023} could reach performances around 80\%. Finally, performances on LobbyMap \citep{morio2023an} are quite low (between 31\% and 57.3\% depending on the strictness of the metric). Overall, the performances show that the datasets are challenging.\footnote{see Table \ref{tab:reported perf stance} in the Appendix for details.}

\paragraph{Exhaustivity} %\fms{misplaced here. } 
Datasets such as ClimateFEVER \citep{diggelmann_climate-fever_2020} and LobbyMap \citep{morio2023an} rely on automated construction methods that may not ensure exhaustivity of evidence coverage. ClimateFEVER utilizes BM25 and Wikipedia for evidence retrieval, which could result in missing relevant information, particularly when compared to more advanced retrieval methods. LobbyMap relies on the \url{LobbyMap.org} website, which was not designed to be exhaustive. These limitations should be further investigated.

\paragraph{Insights} If an organization presents itself as environmentally friendly while simultaneously promoting climate-skeptic narratives or opposing climate remediation efforts, it probably is creating a misleading portrayal of its environmental stance. On the contrary, if they are aligned, it supports the authenticity of the organization's efforts.
Fortunately, stance detection has been extensively studied, yielding promising results; however, current performance levels leave room for improvement. Future research should focus on enhancing both methods and datasets to address these gaps.
While existing datasets provide strong foundations, they may suffer from a lack of exhaustivity, particularly in evidence retrieval and coverage, which requires further investigation. %Addressing these limitations will be critical for advancing the reliability of fact-checking systems in combating greenwashing.
\subsection{Question Answering}
\label{sec:qa}

%Question answering (QA) is a known task:

\task Given an input question and a set of resources (paragraphs or documents), produce an answer to the question. 

\begin{table}[ht!]
\centering
\caption{Summary of datasets, labels, scopes, and sources for climate-related QA datasets.}
\label{tab:qa input}
\begin{tabular}{p{2cm}p{4cm}p{2.5cm}p{3cm}}
\toprule
\textbf{Dataset} & \textbf{Input} & \textbf{Labels}                                                                                  &                                          \\
\midrule
ClimaQA ~\citep{spokoyny2023answering} 
& The text from a \textit{response} to one of the CDP questions; and one of the \textit{questions} from the CDP questionnaire  & \multicolumn{2}{p{6cm}}{\textit{1}: the response answers this question;
\textit{0}: The response does not answer this question, but another one} \\
\midrule
ClimateQA~\citep{luccioni_analyzing_2020} & \textit{Sentence} from reports and a \textit{question} based on the TCFD recommendations & \multicolumn{2}{p{6cm}}{\textit{1}: the sentence answers the question;
\textit{0}: The sentences does not answer the question} \\
\midrule
ClimaINS ~\citep{spokoyny2023answering} 
& The text from a \textit{response} to one of the questions from the NAIC questionnaire & \textit{MANAGE}, \textit{RISK PLAN}, \textit{MITIGATE}, \textit{ENGAGE}, \textit{ASSESS}, \textit{RISKS} & The labels correspond the 8 questions asked in the NAIC questionnaires \\
\bottomrule
\end{tabular}

\end{table}

QA can be used for climate-specific applications such as structuring the information related to climate change from documents  \citep{luccioni_analyzing_2020, tobias_schimanski_climatebert-netzero_2023}, building chatbots with climate-related knowledge to make it more accessible to non-experts  \citep{s_vaghefi_chatclimate_2023, cliamtebot_2022}, or helping identify potentially misleading information  \citep{jingwei_ni_paradigm_2023}. 

\paragraph{Solution} The QA process can be divided into two main steps \citep{krausEnhancingLargeLanguage2023}: the retrieval step and the answer generation step. The retrieval step involves locating the relevant information or answer to the user's query in external documents. The answer generation step involves formulating a response to the user's query from (1) the information retrieved from documents in the first step or (2) the pre-trained internal knowledge of a model. 

\paragraph{Retriever} Retriever systems can be generic, using techniques such as sentence similarity, BM25, or document tags to filter documents and retrieve specific passages. \citet{schimanski-etal-2024-climretrieve} published ClimRetrieve, a benchmark for climate-related information retrieval in corporate reports. They found that simple embedding approaches are limited.
Therefore, several works have specifically focused on improving the identification of answers within climate-related documents \citep{luccioni_analyzing_2020, spokoyny2023answering}. \citet{luccioni_analyzing_2020} reformulated TCFD recommendations as questions and annotated reports to identify answers to the questions. The dataset includes questions paired with potential answers, each labeled to indicate whether the answer addresses the question or not. Based on the QA model trained on this dataset, they also proposed a tool called ClimateQA. \citet{spokoyny2023answering} focused on questions from the CDP questionnaire and the NAIC Climate Risk Disclosure survey. As the responses to those questionnaires are publicly available, they constructed two datasets, ClimaINS and ClimaQA. ClimaQA is similar to ClimateQA (question/potential answer pairs). On the contrary, ClimaINS is a QA dataset framed as a classification task: it contains the responses from the NAIC survey, and each response is labeled as answering one of the 8 questions of the survey.  Once a model has been trained on these datasets, it can be used to search for answers to the questions in other documents (effectively retrieving the information from an unstructured document). 
For ClimaINS, the authors experimented with multiple fine-tuned Transformers. As ClimaQA is framed as a retrieval task, the authors used BM25, sentence-BERT, and ClimateBERT to rank possible answers. Using the methodology evaluated on ClimaQA, they proposed a system to extract responses to CDP questions directly from unstructured reports and automatically fill the questionnaire \citep{spokoyny2023answering}. \citet{luccioni_analyzing_2020} trained and evaluated a RoBERTa-based model called ClimateQA on finding answers to TCFD-based questions. They all reached good performance but with room for improvement,
\footnote{See Table \ref{tab:reported perf question asnwering (qa)} in the Appendix for details.} demonstrating the feasibility of the task but also the need for further research.

\paragraph{Answer generation} After the retrieval, the second step of question answering is to generate an answer for the question (given input resources or not). The generation can be simple such as finding a particular information in a paragraph. For example, \citet{tobias_schimanski_climatebert-netzero_2023} used a QA model (RoBERTa SQuaD v2 \citep{rajpurkar-etal-2016-squad}) to extract the target year, the percentage of reduction, and the baseline year from the reduction target of a company from an input text that contains an emission reduction target. While it is possible to rely on generalist models, as they are proficient few-shot learners \citep{lm_few_shot_learner, thulke2024climategpt}, multiple studies proposed using domain-specific models.  
Climate Bot \citep{cliamtebot_2022} built a dataset of climate-specific QA. Given a question and a document (scientific/news), the model should find the span answering the question.  \citet{mullappilly-etal-2023-arabic} proposed a dataset of question-answer pairs based on ClimaBench \citep{spokoyny2023answering} and CCMRC \citep{cliamtebot_2022} to train a model to generate answers. \citet{thulke2024climategpt} introduced a climate-specific corpus of prompt/completion pairs for Instruction Fine-Tuning created by experts and non-experts. 
\citet{cliamtebot_2022} finetuned an ALBERT model on their climate-specific QA dataset. \citet{mullappilly-etal-2023-arabic} trained a Vicuna-7B on their climate-specific answer generation dataset. And \citet{thulke2024climategpt} trained LLama-2-based models (ClimateGPT) on their domain-specific prompt/completion dataset. They also evaluated their ClimateGPT model alongside multiple LLMs on climate-related benchmarks.\footnote{They found that the domain-specific models outperformed generalist ones (see Table \ref{tab:reported perf question asnwering (qa)} in the Appendix for details).}

\paragraph{Retrieval Augmented Generation} By combining the last two components (a retriever and a QA system), one can develop Retrieval-Augmented Generation (RAG) systems. \citet{cliamtebot_2022} provide the first example of a RAG system in climate-related tasks based on sentence-BERT to retrieve documents and AlBERT to identify the answer. More recently, \citet{jingwei_ni_paradigm_2023}~proposed ChatReport, a methodology based on ChatGPT for analyzing corporate reports through the lens of TCFD questions using RAG.  The reports, the answers to the TCFD questions, and a TCFD conformity assessment are chunked and stored in a vector store for easy retrieval. Each chunk and/or answer can be retrieved and injected into the prompt. They added in the prompt the notions of greenwashing and cheap talk to invite the model to provide a critical analysis of the retrieved answers. Similarly, \citet{s_vaghefi_chatclimate_2023} introduced ChatClimate, a RAG-based pipeline to augment GPT-4 with knowledge about IPCC reports. While the previous LLM-based approaches rely on closed-source models, \citet{mullappilly-etal-2023-arabic} and \citet{thulke2024climategpt} proposed RAG-based pipelines relying on open-source models.
\citet{cliamtebot_2022} is the only study that evaluated their RAG pipeline using classical metrics (F1-score, BLEU, METEOR). Unfortunately, those metrics are not well aligned with human judgment for text generation \citet{chen-etal-2019-evaluating}. Therefore \citet{s_vaghefi_chatclimate_2023} and \citet{mullappilly-etal-2023-arabic} relied on human or ChatGPT evaluations. They conclude that their approaches improve on non-domain-specific RAG systems.

\paragraph{Insights} The main application of Question-Answering (QA) systems is to analyze complex documents, such as corporate climate reports, that can be particularly useful for greenwashing detection. Modern LLMs achieve strong results in QA due to instruction-following fine-tuning, which improves adherence to specific instructions, and Retrieval-Augmented Generation (RAG), which bases its answers on retrieved documents. However, retrieval systems  remain a performance bottleneck \citep{maekawa-etal-2024-retrieval}. Furthermore, answer generation is challenging to evaluate \citep{survey_nlg_eval}, often requiring human feedback or advanced model-based assessments.

\subsection{Classification of Deceptive Techniques}
\label{sec:deceptive}

In analyzing climate-related discourse, it can be useful to identify rhetorical strategies that might obscure or misrepresent an entity’s stance. 
For example, expressing support for climate remediation policies on one side \citet{morio2023an}, and promoting arguments downplaying the urgency of climate change on the other \citet{coanComputerassistedClassificationContrarian2021} could signal a lack of authenticity. Another example could be promoting a product through misleading rhetoric, by, for example, claiming that a product is better because it is natural, which would be an ``appeal to nature'' -- a fallacious argument \citet{vaid-etal-2022-towards, jain_supply_2023}.
Detecting these deceptive techniques could help identify misleading communications.

\begin{table}[ht]
\centering
\caption{Summary of datasets, labels, scopes, and sources for tasks related to deceptive techniques in climate-related context.}
\label{tab:datasets_deceptive}
\begin{tabular}{p{2.5cm}p{3cm}p{3cm}p{3cm}}
\hline
\textbf{Dataset}  & \textbf{Input}  & \textbf{Labels}          & \textbf{(Labels details)}                                        \\
\hline
LogicClimate \citep{jin-etal-2022-logical}  & texts from climatefeedback.org  & \textit{Faulty Generalization, Ad Hominem, Ad Populum, False Causality, ...} & Classifies fallacies (Multi-label)\\
\hline
\raggedright Neutralization Techniques \citep{bhatia_automatic_2021-1} & paragraphs from other previous works on climate-change & \textit{Denial of Responsibility, Denial of Injury, Denial of Victim, Condemnation of the Condemner, ...} & Classifies neutralization techniques  \\
\hline
Contrarian Claims \citep{coanComputerassistedClassificationContrarian2021} & paragraphs from conservative think tank & \raggedright \textit{No Claim, Global Warming is not happening, Climate Solutions won't work, Climate impacts are not bad, ...} &  Classifies arguments into  super/ sub-categories of climate science denier arguments  \\
\hline
\end{tabular}
\end{table}

\task The goal is to classify statements into argumentative categories: fallacies, types of arguments, or rhetorical techniques.

\zeroshot \citet{divinus_oppong-tawiah_corporate_2023} frame greenwashing as fake news. They propose to tackle this identification through the form perspective. 
They developed a profile-deviation-based method to detect greenwashing in corporate tweets by comparing linguistic cues (e.g., quantity, specificity, complexity, diversity, hedging, affect, and vividness) to theoretically ideal profiles of truthful and deceptive communication. They compute a greenwashing score as Euclidean distances.

\datasets An organization seeking to justify limited action might adopt a rhetoric that downplays the urgency of global warming or dismisses the impact of negligent behavior. These kinds of climate contrarian arguments mostly fall into a finite set of categories (e.g. ``Solutions will not work'', or ``Human influence is not demonstrated''). Therefore, \citet{coanComputerassistedClassificationContrarian2021} constructed a taxonomy of such contrarian claims and published a large dataset annotated with such claims. 
Similarly, \citet{bhatia_automatic_2021-1} proposed a dataset to classify neutralization techniques, i.e., rationalizations that individuals use to justify deviant or unethical behavior (e.g. Denial of the victim). Finally, \citet{jin-etal-2022-logical} studied fallacious arguments and how they apply to climate change more broadly. They constructed LogicClimate, a climate-specific dataset of sentences annotated with fallacies, using articles from the \href{https://science.feedback.org/reviews/?_topic=climate}{Climate Feedback website}.

\solutions Fine-tuned Transformer architectures and classical approaches have been evaluated on each dataset \citep{jin-etal-2022-logical, bhatia_automatic_2021-1, coanComputerassistedClassificationContrarian2021}. \citet{jin-etal-2022-logical} experimented with an ELECTRA model, fine-tuned to use the structure of the argument. The performance of fine-tuned approaches ranges from 58.77\% to 79\%. The models struggle with fallacy detection in LogicClimate, but perform well with generalist fallacy detection, contrarian claims, and neutralization techniques. \citet{thulke2024climategpt} experimented with multiple zero-shot approaches using multiple LLM on the binary classification of contrarian claims using \citet{coanComputerassistedClassificationContrarian2021}'s dataset\footnote{
    Their climateGPT-70B and LLama-2-Chat-70B models both reached an F1-score of 72.5\% (see Table in the Appendix for details).}.

\paragraph{Insights} The low performance of models on tasks such as fallacy detection and neutralization classification indicates the inherent complexity of these tasks. Fallacy detection, for one, is known to be inherently subjective \citep{helwe-etal-2024-mafalda}.
The detection of neutralization techniques, too, appears to be subjective, as even human annotators achieve only a moderate performance level of 70\% F1-score.
Furthermore, both LogicClimate \citep{jin-etal-2022-logical} and the neutralization dataset \citep{bhatia_automatic_2021-1} are very small in size.
To address these limitations, future research could explore several directions, including increasing the size of the datasets, defining  the labels more precisely,
or permitting multiple correct annotations \citep{helwe-etal-2024-mafalda} to acknowledge viable disagreement. Additionally, the analyses on rhetorical techniques could be combined with other analyses (such as stance detection) to search for communication patterns. 

\subsection{Environmental Performance Prediction}
\label{sec:env prediction}

Greenwashing can be interpreted as a misrepresentation of the company or product's environmental performance. The environmental performance is often summarized by a quantitative metric such as the ESG score, the Finch score\footnote{The Finch Score is a sustainability rating system designed to help consumers make eco-friendly choices by evaluating products on a scale from 1 to 10. See \url{https://www.choosefinch.com/}}, or CO2 emissions. These scores are usually not directly mentioned in the company reports, but have to be inferred based on company communications:

\task Given an input company report, output an environment-related quantitative value (such as the ESG score or the amount of Carbon Emission), even if that value is not mentioned in the report. In variants of this task, the goal is to predict not a value, but a range for that value, out of a set of possible ranges.

\zeroshot \citet{jain_supply_2023} proposed to predict the Scope 3 emissions 
from the list of financial transactions of the company. They first performed classification of all transaction descriptions to map them to their emission factors (emission per \$ spent), and then compute the emissions of each transaction. For ESG score prediction, \citet{bronzini_glitter_2023} proposes to use LLMs as few-shot learners to extract triples of ESG category, action, and company from sustainable reports, and to construct a graph representation with few-shot learning. They demonstrated the triplet generation using Alpaca, WizardLM, Flan-T5, and ChatGPT on a few examples. They could analyze disclosure choices and company similarities using the constructed graph. More importantly, they also used the graph to interpret ESG scores through the interpretability analysis of OLS predictions, effectively predicting the ESG score. 

\datasets \citet{Mehra_2022} proposed to predict the changes in the ESG score instead of the actual value. They constructed their dataset by extracting the three sentences most relevant to the environment from financial reports and associated them with the E score\footnote{ESG scores are usually aggregated scores along multiple dimensions. In this study, they are focusing on the "Environment" part of the ESG score.}'s change and direction of change. Instead of relying on scores, \citet{clarkson_nlp_us_csr} proposed to focus on good/bad CSR (corporate social responsibility) performers. Their approach evaluates CSR performance based on linguistic style rather than content, aiming to identify whether language patterns alone influence the perception of CSR quality. \citet{Greenscreen} introduced a multi-modal dataset of the tweets of companies to predict a company's ESG unmanaged risk. 
Finally, focusing on products and not companies, \citet{linSUSTAINABLESIGNALSijcai2023} introduced a dataset of online product descriptions and reviews used to predict the Finch Score. All these scores can be used to identify companies and products that are actually environmentally friendly, helping users distinguish actual sustainability from greenwashing. However, they can also be used to find discrepancies between the communicated green-ness of a product or company (e.g. the percentage of climate-related texts in reports) and the actual green-ness (e.g. ESG score or quantity of emissions), which might indicate cases of greenwashing.  

\solutions \citet{Mehra_2022} evaluated fine-tuned Transformers on ESG change prediction. \citet{clarkson_nlp_us_csr} experimented with hand-chosen features plugged into random forest and SVM on CSR performers prediction. \citet{linSUSTAINABLESIGNALSijcai2023} also experimented with traditional approaches (e.g. gradient boosting) and custom architectures based on Transformer models on Finch score prediction. \citet{Greenscreen} introduced baseline models, which are simple image and text embedding models (e.g., CLIP or sentence-BERT) with a classification head on ESG score prediction.
\citet{bronzini_glitter_2023, clarkson_nlp_us_csr, Greenscreen} did not report baselines, which makes the study difficult to analyze. \citet{Mehra_2022} reported good accuracy, demonstrating that a few sentences hold a significant amount of information to predict ESG score changes. However, there is definitely room for improvement, indicating that the information selected is not sufficient. \citet{lin-etal-2023-linear} reported the performance of the baseline (average score) reaching already mean squared error (MSE) of 11.7\% already quite low. Their Transformer-based approach reached MSE of 7.4\% improving slightly on classical approaches (Gradient Boosting reaching MSE of 8.2\%)\footnote{
Detailed performances can be found in Table \ref{tab:appendix env pred} in the Appendix.}.

\paragraph{Insights} We expected the prediction of ESG and CSR metrics to be difficult and require an extensive understanding of a company. However, \citet{bronzini_glitter_2023, linSUSTAINABLESIGNALSijcai2023, Mehra_2022} and \citet{clarkson_nlp_us_csr} showed that it is possible to build strong predictors relying only on textual elements. This can be explained because analysts reward transparency \citep{bronzini_glitter_2023}, so the quantity and complexity of disclosure have strong predictive power \citep{clarkson_nlp_us_csr}. Although existing studies have examined specific dimensions, future research should investigate the interaction between form \citep{clarkson_nlp_us_csr} and content \citep{bronzini_glitter_2023}, as this could help uncover inconsistencies that may indicate greenwashing practices.

\section{Greenwashing Detection}
\label{sec: greenwashing signals}

After having discussed different subtasks of greenwashing detection, we now come to the final, all-englobing task:

\task Greenwashing detection is the task of predicting if a text contains greenwashing. This is a binary classification task. Greenwashing detection can also be done at an aggregated scale (such as by a yearly indicator).

\paragraph{Greenhushing and Selective Disclosure}  There is a large body of work on disclosure (as described in Section~\ref{sec: tcfd}). Quantifying climate-related disclosure can highlight greenhushing and selective disclosure. \citet{bingler_cheap_2021} identify a lack of disclosure in the Strategy and Metrics\&Targets categories, which they describe as \textit{cherry-picking}. They also highlight that merely announcing support for the TCFD does not lead to an increase in disclosure -- on the contrary, it is a practice called \textit{cheap talk}. Similarly, \citet{auzepy_evaluating_2023} conducted a more fine-grained analysis of TCFD-related disclosure in the banking industry. They also highlighted large differences across TCFD categories. In particular, they found a low disclosure rate on the fossil-fuel industry-related topic despite large investments in that sector.

%(omission type greenwashing \citet{defreitasnettoConceptsFormsGreenwashing2020a}).

\paragraph{Climate Communication as an Image-Building Strategy} \citet{dingCarbonEmissionsTCFD2023} and \citet{chou_ESG} found a correlation between climate-related disclosure and carbon emission: companies that emit more tend to disclose more climate-related information. \citet{bingler2023cheaptalkspecificitysentiment} proposed a Cheap talk index based on claims specificity. They reached the same conclusion that larger emitters tend to disclose more, but also that negative news coverage is correlated with cheap talk. \citet{marco_polignano_nlp_2022} showed that while disclosing more, reports mostly focus on positive disclosure.  \citet{hyewon_kang_analyzing_2022} proposed a sentiment ratio metric that highlights overly positive corporate reports. They identified periods of overly positive communications that followed negative environmental controversies; in other words, companies that tried to rebuild their image after a controversy. 
\citet{csr_report_greenwashing} analyzed CSR reports of environmental violators and companies with clean records. They studied an ensemble of variables: quantification of environmental content, readability score, and sentiment analysis. They found that violators publish longer, more positive, and less readable CSR reports with more environment-related content.
This suggests that companies use climate disclosure as a tool to mitigate controversies and/or to improve their image. \citet{kdir23} propose to use the discrepancies between internal disclosure and social media perception of a company to identify potential greenwashing.

\paragraph{The Role of Disclosure Style in Perceived Commitment} \citet{clarkson_nlp_us_csr} found that companies that use a more complex language in their CSR disclosure are associated with a higher CSR rating. \citet{schimanski_bridging_2023} concluded that a higher quantity of ESG communication is associated with higher ESG rating. \citet{rouenEvolutionESGReports2023} also highlighted the relationship between disclosure quantity and complexity with the ESG score. This might indicate that the linguistic style is a good predictor of the substance of the discourse, or that analysts are rewarding companies that communicate on ESG-related issues.

\paragraph{ESG Score in Greenwashing Detection} \citet{LEE_greenwashing} proposed a greenwashing index based on the difference between the ESG score and the ESG score, weighted by the quantity of communication on each topic (E, S and G). 
%The quantity of communication is computed using dictionaries. 
\citet{Greenscreen} define two types of risks: managed risk (the company is addressing it), and unmanaged risk (the company is not currently addressing it but it could). Based on these, they propose an ESG unmanaged risk score as a measure for environmental performance, which can then be used as a signal for identifying greenwashing.

\paragraph{Contrasting Stances: A Method to Identify Greenwashing in Climate Communications} \citet{morio2023an} proposed to evaluate a company's stance %polarity 
towards climate change mitigation policies. They hypothesize that companies with a mixed stance might be engaging in greenwashing. The hypothesis was not tested and remains unconfirmed; still, such methodology helps understand the narratives of companies around climate change. \citet{coanComputerassistedClassificationContrarian2021} were able to analyze the type of contrarian claims used by conservative think-tank (CTTs) websites and contrarian blogs. They showed a shift over time from climate change denying arguments to opposing climate change mitigation policies. Despite this shift, they found that the primary donors of these think tanks continue to support organizations promoting a climate-denier narrative aimed at discrediting scientific evidence and scientists. Discrepancies between the stance extracted from official communications~\citep{morio2023an} on the one hand, and the stance of third-party media organizations such as think tanks founded by the company \citep{coanComputerassistedClassificationContrarian2021} on the other, might indicate greenwashing.

\paragraph{Defining the linguistic features of greenwashing} \citet{divinus_oppong-tawiah_corporate_2023} we propose to identify greenwashing using the linguistic profile of tweets. They estimate the deceptiveness of the text using a keyword-based approach and linguistic indicators (e.g. word quantity, sentence quantity). They found a correlation between greenwashing and lower financial performances. A scoring system based solely on linguistic elements has significant limitations. For example, the following tweet gets a  high greenwashing score: ``Read about \#[company’s] commitment to a \#lowcarbon future http://[company website]''. However, it is merely an invitation to read a Web page. 

\paragraph{Insights} Several indicators for greenwashing have been proposed: 
overly positive sentiment \citep{hyewon_kang_analyzing_2022}, the use of specific linguistic cues~\citep{divinus_oppong-tawiah_corporate_2023}, discrepancies between ESG scores and corporate disclosures~\citep{LEE_greenwashing, Greenscreen}, ambiguous stances~\citep{morio2023an}, and inconsistencies between social media perception and official disclosures~\citep{kdir23}.
While these approaches laid the theoretical groundwork for understanding indicators of greenwashing, they have a significant limitation: they are not empirically evaluated~\citep{measuring_greenwashing}. 
Only a few studies, such as that by \citet{hyewon_kang_analyzing_2022}, have made attempts to validate their signals against actual cases, but even these efforts have not resulted in a comprehensive dataset of greenwashing examples.
This gap between theory and practice highlights the necessity of developing datasets containing real-world instances of greenwashing. Without empirical evaluation, the prediction that a company engages in greenwashing is unfounded at best, and misleading or even defamatory at worst.  
However, building such a dataset comes with many challenges. First, one would have to identify suitable sources of documents that are openly accessible and likely to contain greenwashing. Second, one would have to overcome the inconspicuous and subjective nature of greenwashing itself. Finally, the publication of any such dataset exposes the authors to charges of defamation by the companies they accuse of greenwashing.

\section{Open Challenges in Greenwashing Detection}
\label{sec:challenges}

While the lack of real-world greenwashing datasets constitutes a major obstacle, it is only one of several unresolved challenges. Building on the reviewed literature, this section examines additional limitations that affect greenwashing detection and, more broadly, NLP tasks characterized by ambiguity, subjectivity, and real-world constraints. Although discussed through the lens of greenwashing-related applications, these challenges are not specific to this domain and reflect broader methodological issues in applied NLP. We focus on three interconnected dimensions: (\ref{sec:eval difficulty}) evaluation methodology, where ambiguity and task complexity complicate performance assessment; (\ref{sec:robustness}) model robustness, including sensitivity to noise and data imbalance; and (\ref{sec:data}) data sources, encompassing reproducibility, dataset availability, and definitional issues.

\subsection{Evaluation Methodology}
\label{sec:eval difficulty}

Firstly, as models improve, they are able to tackle more complex tasks, moving from classification to text generation, and from extraction to reasoning. This progression in task complexity creates an increase in evaluation complexity. \smallskip

\paragraph{Classification Evaluation} Many tasks in greenwashing detection are framed as classification tasks. Such a framing may be too simplistic, in that it is blind to the semantic similarity between the classes. For example, in the Contrarian Claims dataset~ \citep{coanComputerassistedClassificationContrarian2021}, models may fail on the fine-grained classification, but still get the super-category right. %it is likely that the correct and the wrong labels that are part of the same super-category.
%\oana{why do you use "while" here? so they fail on both super category and sub category, why is this surprising?}\tom{Same super category not =/= categories}. I see, but it seems intuitive, the classes are probably more similar when in the same super category.
For such cases, \citet{morio2023an} proposed  strict and relaxed metrics, where the relaxed metric grouped similar labels together (e.g. "strongly supporting" and "supporting" grouped in "supporting" label). 
%This helps understand how the models actually perform.% and mitigates the weakness of the datasets. 
%We believe it is essential to provide more thorough metrics to better evaluate AI alignment with human evaluation.

\paragraph{Ambiguous Labels} One particular issue is that the main hypothesis of the classification task is that categories are distinguishable from each other. This is not always the case. 
For example, specificity classification  \citep{bingler2023cheaptalkspecificitysentiment} is the task of distinguishing specific claims from unspecific claims. It turns out that this task is itself too unspecific, % Fabian: small joke, feel free to remove...
as the low IAA shows. While this issue appears obvious for this particularly ambiguous task, other less ambiguous classes also contain an ambiguous frontier. For example, climate/non-climate labels can be debated in some instance: \textit{"The company produced 10 million computers this year}" or "\textit{The company produced 10 million solar panels this year}" are respectively predicted "non-climate" and "climate" by climateBERT's climate detection model  \citep{bingler2023cheaptalkspecificitysentiment}. In this example, the sentences are fundamentally similar. If the company's business model is to produce solar panels, then this statement is not necessarily about climate, at least not more than the other one.

\paragraph{Generated Text Evaluation} Some tasks are not classification tasks, but text generation tasks. Evaluating generated text is a challenging task~ \citep{survey_nlg_eval}. We argue that evaluating generated text is still a challenge deserving future research. 
In the papers we considered for this review, to evaluate generative models, we came across solutions such as using BLEU, METEOR, human-based evaluation, and LLM-based evaluation~ \citep{mullappilly-etal-2023-arabic, cliamtebot_2022, s_vaghefi_chatclimate_2023, thulke2024climategpt}. BLEU and METEOR are not great to evaluate tasks such as question answering \citep{chen-etal-2019-evaluating}. 
Human evaluation also has many shortcomings. Firstly, it is not possible to perform an automatic evaluation; the evaluations are therefore limited to smaller sizes and variations in settings. Secondly, human evaluation of generated texts often relies on comparisons  \citep{mullappilly-etal-2023-arabic, thulke2024climategpt}. While \citet{thulke2024climategpt} refers to specific comparison guidelines  \citep{bulian2024assessinglargelanguagemodels},  \citep{mullappilly-etal-2023-arabic} relies only on the evaluator's preferences.
Thirdly, human evaluation can be influenced by factors other than the quality of the answer -- for example, the observer-expectancy effect  \citep{rosenthal1976}.
For example, in ChatClimate~ \citep{s_vaghefi_chatclimate_2023}, humans evaluate the accuracy of text generated by a generic ChatGPT vs. the proposed ChatClimate chatbot. The text produced by the ChatClimate bot always gives specific references (down to page numbers), while the generic ChatGPT does not. It is thus very easy for evaluators to identify the more accurate answer simply by looking at the type of references, so that there is the danger that this was in fact the only facet of the texts that was evaluated (and not, say, the general accuracy or quality of the answer). 
A third shortcoming of evaluations appears with LLM-based evaluation. Language models are known to reproduce human biases~ \citep{gallegos2024biasfairnesslargelanguage}: they are sensitive to the way questions are formulated, to the order of answers in multiple-choice questions~ \citep{pezeshkpour-hruschka-2024-large}, and to cues that can influence human evaluation such as using ``Our Model''/``Their model'' (which is the wording used in \citet{mullappilly-etal-2023-arabic}).
LLM-based evaluation should thus always adhere to the same rigorous standards that are applied to human evaluation.
Proper control of the evaluation setting is crucial. Subtle factors, such as the specific wording used or the unintentional leakage of task-related information, can bias the results. Therefore, strict protocols and a meticulously designed evaluation environment are essential to mitigate these risks.

\begin{figure}
    \centering
    \includegraphics[width=0.85\linewidth]{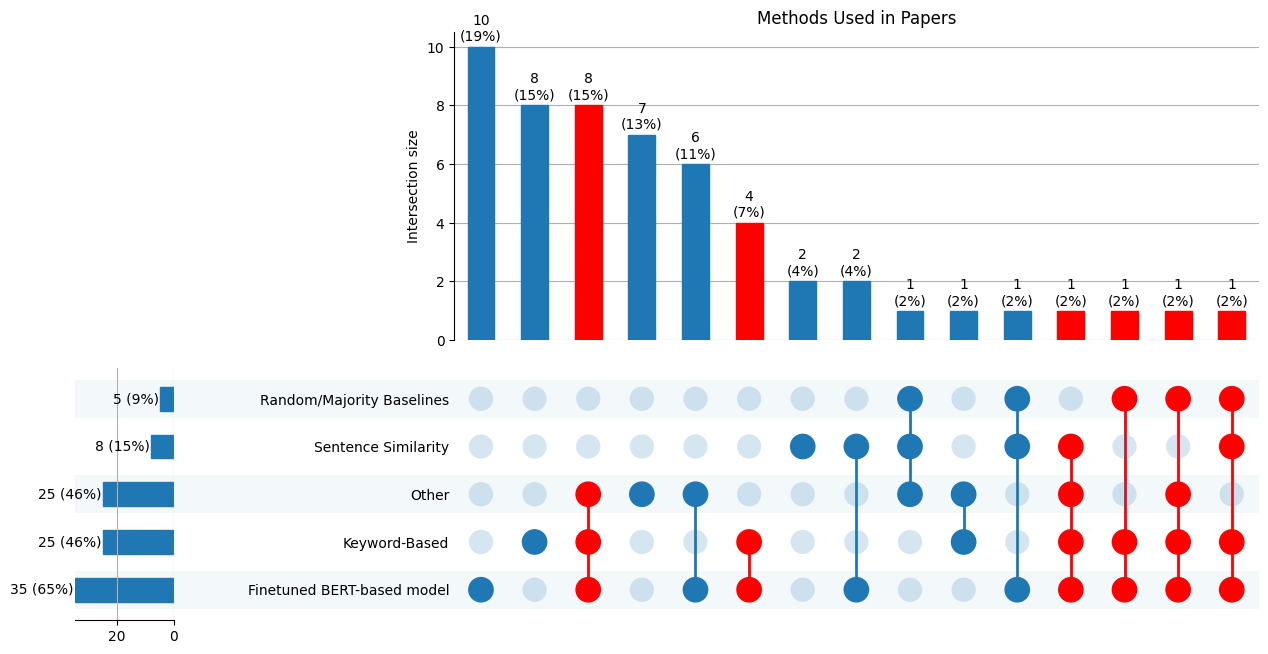}
    \caption{Upset plot \citep{upset_plot} of the methods used in the studies mentioned in this literature review. Each bar counts the number of study reporting a given set of methods. The sum of a paper using a given method is reported on the horizontal bars. We have highlighted in red the intersection sets that include both fine-tuned Transformers and keyword-based methods.}
    \label{fig:models}
\end{figure}

\paragraph{Baselines} Figure~\ref{fig:models} shows the approaches that are commonly used in greenwashing detection tasks. We see that the most common solution is fine-tuning transformer models; more recent approaches include using LLMs as few-shot learners. However, when evaluating, the state-of-the-art models are compared to baselines. We argue that some basic baselines are often forgotten, even though they provide important insights:
\begin{itemize}
    \item \textit{Random/Majority Baseline:} Any proposed approach should beat a random baseline. However, we find that only 9\% of papers we studied compare their results to the random or majority baseline. For example, \citet{Friederich_climate_risk_disclosure} reported performances under 50\% for a binary classification task, which is worse than a theoretical random classifier on a balanced dataset. Their dataset is not balanced, but it is difficult to draw conclusions without the performance of a random classifier. 
    Another interesting aspect of reporting the random classifier performance is to have a point of reference, or a lower baseline. For example, on SciDCC or ClimaTOPIC, the performance of models is around 50-60\%, yet this is excellent performances as the random classifier performed around 5-7\%  \citep{spokoyny2023answering}. In case of an heavily imbalanced dataset, providing the performance of the majority baseline might be more meaningful. If the dataset is not avaiable, it is not possible to estimate that value, which is the case for  \citep{Friederich_climate_risk_disclosure}. Similarly a constant baseline help understand the performance of a regression model. \citet{Greenscreen} and \citet{bronzini_glitter_2023} evaluated the prediction of ESG score, yet they do not provide baseline. If the values are really concentrated, a model which always predict the mean value might have a good performance. Unfortunately, without the distribution of values, it is not possible to estimate it. 
    \item \textit{Simple Baselines:} If a keyword-based baseline performs well, or even better than a proposed method, then the proposed method is likely an overkill. And yet, less than half of the works we studied report the performance of a keyword-based baseline.
    Studies such as \citet{rouenEvolutionESGReports2023, clarkson_nlp_us_csr}, which showed that linguistic features such as the quantity and complexity of text are good predictors for the ESG score, therefore seemingly simple predictors (e.g. a keyword-based classifier) might perform surprisingly well, even on complex tasks. Moreover, we observed that the simple baseline can even outperform finetuned approaches, for example on TCFD classification  \citep{sampson_tcfd-nlp_nodate} or on ClimaINS dataset  \citep{spokoyny2023answering}.
    Beyond a validation (or invalidation) of the proposed approaches, a keyword-based baseline would also allow us to understand if models rely on an ``understanding'' of the text or merely on word cues. 
    \item \textit{Human Baselines:} Using a human as a baseline can help determine whether a task is well-defined. If even a qualified human cannot achieve high accuracy, this most likely means that the task is subjective, ill-defined, or in disagreement with the gold standard. Vice versa, if a human achieves high accuracy, but the model achieves even higher accuracy, this could mean that the model has acquired expert knowledge. However, most of the studies discussed here do not incorporate an evaluation of human performance (Only 3 studies reported Human performances).
\end{itemize}

\subsection{Model Robustness}
\label{sec:robustness}

When evaluating a model, it is important to know whether the performance on a dataset is due to an inherent capacity of the model or due to random factors that may be entirely different on another dataset.

\paragraph{Model Performance Evaluation} Many factors can influence the performance of a model (e.g. data sampling, hyper-parameters, random initialization, model architecture). Moreover, we see that many studies report differences in performance of less than 1\%. This difference might be due to the choice of model, but it could also be due to the other factors. Quantifying the uncertainty induced by those factors helps understand if a difference is significant. This could be done using confidence intervals, p-values or simply reporting the standard deviation.
However, we found only seven papers that quantify the uncertainty of the model performance~ \citep{nicolas_webersinke_climatebert_2021,varini_climatext_2020,bingler2023cheaptalkspecificitysentiment,Schimanski2024nature,schimanski_bridging_2023,LEE2023119726,tobias_schimanski_climatebert-netzero_2023}. 

\paragraph{Adversarial Attacks and Noisy Text} While fine-tuning leads to the best performances, it is also prone to word-based adversarial attacks~ \citep{vinicius_woloszyn_towards_2021}, as the training data are often relatively small, and might not include syntactically diverse examples. 
Noisy texts are an inherent part of real-world applications. Yet, most studies mentioned in this review trained and evaluated models on clean datasets (containing only clean examples) and  did not quantify the robustness of their approaches to noise. 

\paragraph{Imbalanced Datasets} In most of the tasks, the class of interest is relatively rare. For example, only 5\% of text is in the positive class ``climate-related'' in the study of~\citet{yu_climatebug_2024}. This means that (1) a simple baseline that always says ``non-climate-related'' will have a very high precision, and (2) a model trained on such a dataset would probably overfit. Therefore, most studies mentioned in this review trained and evaluated the models on relatively balanced datasets. 
On the contrary, a model that was evaluated only on a balanced dataset may fail spectacularly when tested on real-world, unbalanced datasets~ \citep{Friederich_climate_risk_disclosure}. For example, a model reaching an F1-score of 90\% on a balanced dataset would significantly drop on an unbalanced dataset with 5\% of text in the positive class (to a binary F1-score of 47\%, and a macro F1-score of 71\%). This is due to the large number of false positives.
For tasks related to greenwashing detection, annotating data is a time-consuming and resource-intensive process, often resulting in an imbalanced dataset that fails to represent all relevant cases adequately. This imbalance complicates the model's performance trade-off: a higher recall is desirable when providing leads for human analysis to avoid missing potential greenwashing cases. However, if the goal is to accurately measure greenwashing prevalence, minimizing false positives is critical, as false accusations can have serious repercussions.

\subsection{Data Sources}
\label{sec:data}

\paragraph{Close-source data and reproducibility} The studies mentioned in our review use various data sources, most of which are publicly available (annual and sustainable reports used in 18 studies)%~ \citep{doran_risk_disclosure, kolbel_ask_2021, chou_ESG, huangFinBERTLargeLanguage2020, krausEnhancingLargeLanguage2023, kheradmand2021a, rouenEvolutionESGReports2023, lai_using_2023, hyewon_kang_analyzing_2022, auzepy_evaluating_2023, bjarne_brie_mandatory_2022, nicolas_webersinke_climatebert_2021, luccioni_analyzing_2020, kheradmand2021a, bronzini_glitter_2023, LEE2023119726, marco_polignano_nlp_2022, krausEnhancingLargeLanguage2023}, 
social media (6 studies)%~ \citep{vinicius_woloszyn_towards_2021, luo_detecting_2020, vaid-etal-2022-towards, coanComputerassistedClassificationContrarian2021, divinus_oppong-tawiah_corporate_2023, Greenscreen}, 
specialized website (5 studies)%~ \citep{morio2023an, tobias_schimanski_climatebert-netzero_2023, jin-etal-2022-logical, varini_climatext_2020, coanComputerassistedClassificationContrarian2021})
. However, multiple works rely on closed-source data, this include ESG score, carbon emissions, machine-readable reports that are collected through \textit{private providers} such as Refinitiv, Reuters, RavenPack, Trucost, Bloomberg, RobecoSAM, Compustat or Wharton (17 studies)%~ \citep{SAUTNER_cliamte_change_exp, bingler2023cheaptalkspecificitysentiment, stammbach_environmental_2023, Friederich_climate_risk_disclosure, bjarne_brie_mandatory_2022, LEE_greenwashing, nicolas_webersinke_climatebert_2021, liCorporateClimateRisk2020, clarkson_nlp_us_csr, liCorporateClimateRisk2020, SAUTNER_cliamte_change_exp, Mehra_2022, Greenscreen, clarkson_nlp_us_csr, schimanski_bridging_2023, avalon_vinella_leveraging_2023, krausEnhancingLargeLanguage2023}
. Relying on private providers limits the reproducibility of those works, for both academic work and applications.

\paragraph{No greenwashing datasets} While theoretical frameworks have provided a starting point for defining greenwashing, a critical gap remains: the lack of a comprehensive, real-world dataset. This can be explained by several factors: 
\begin{itemize}
    \item Greenwashing definitions are often vague and ambiguous, leaving room for interpretation, making it difficult to use for annotating examples of greenwashing
    \item Green claims are often not self-contained, meaning that the full content of information required to judge if a claim is misleading or not is spread: across multiple paragraphs in a same document, or even across multiple documents. Therefore collecting all the elements required to correctly identify greenwashing is a challenges on its own.
    \item Understanding the claim and annotating greenwashing requires expertise, and sometimes debates to make sure the communication is misleading, making the annotation task even harder.
    \item Greenwashing is a strong accusation, that might impact a company reputation, and that can also, if not substantiated with evidence, be considered as defamation.
\end{itemize}
To address these challenges, several potential data collection strategies can be considered. To avoid relying on a subjective annotation, researchers can rely on a third-party judgement. For example, using news articles that explicitly discuss greenwashing, drawing on prior research focused on detecting media mentions of greenwashing~ \citep{gourierGreenwashingIndex2024}. Another valuable resource could include regulatory records from authorities such as the Advertising Standards Authority (ASA) or databases like \url{greenwash.com}.
Alternatively, previous works tried to produce signals that could potentially indicate greenwashing, future research might rely on such indicators to first filter-out texts that are potentially containing greenwashing, and annotate those selected texts. Intermediary tasks are also important for other reasons. First, decomposing tasks often results in better performance, as supported by the principles behind Chain-of-Thought. Second, when AI systems are designed to include human oversight, it is crucial for these intermediary tasks to remain understandable and accessible to users. Thirdly, the design of energy-efficient AI systems can benefit from employing simpler models for tasks where high precision is not required, such as document retrieval for RAG. Finally, relying on intermediary tasks for the evaluation of models' performances is necessary for both understanding better the models, their limits, and to propose auditable pipelines with humans in the loop.  

\paragraph{Relying on regulatory texts} As mentioned above, two of the challenges in building a dataset of examples of greenwashing are using definitions that are subjective and mitigating risks of defamation. Using regulatory texts as reference helps mitigate those two issues. Firstly, it can help build a proper definition of the label. For example, \citet{stammbach_environmental_2023} rely on the definition of green claims by the European Commission to build their definition of environmental claims. Secondly, instead of emitting a judgment, the method can link a statement to a regulatory text. For example, all works focusing on TCFD simply link a statement to a recommendation  \citep{dingCarbonEmissionsTCFD2023, auzepy_evaluating_2023, bjarne_brie_mandatory_2022, bingler_cheap_2021, sampson_tcfd-nlp_nodate}. Recently, the European Commission adopted the CSRD which defines more precisely greenwashing. Future research could start by finding content related to parts of the regulatory texts related to greenwashing. 

\paragraph{Quantifying communication vs Quantifying information} With the exception of \citet{bronzini_glitter_2023}, the studies reviewed in this work primarily focus on quantifying textual content related to climate, TCFD disclosures, or specific topics of interest. However, as highlighted by \citet{bingler2023cheaptalkspecificitysentiment}, companies increasingly engage in "cheap talk," where statements lack substantive commitments or actionable details. This underscores the need to assess not only the volume of climate-related communication but also the informational content conveyed. For instance, a company might repeatedly emphasize its objectives or highlight a single initiative, creating the illusion of widespread organizational change, while failing to implement meaningful action. Future research should focus on quantifying the information instead of the amount of text related to a topic.

\paragraph{Restricting the definition of greenwashing} Except for obvious lies, what counts as ``misleading communication'' is intrinsically subjective. For some, disproportionately highlighting a company’s green initiatives is misleading (\citet{LEE_greenwashing}); others interpret the same behavior as employees’ pride in early achievements or as a strategic shift to engage stakeholders around more ambitious climate action. Likewise, \citet{hyewon_kang_analyzing_2022} frame greenwashing as overly positive texts, yet such positivity can be a stylistic choice rather than deceptive intent. Moving from a broad to a narrower definition can improve the objectivity of what is considered greenwashing. The trade-off with narrower definitions is twofold: (i) they may exclude other forms of greenwashing, and (ii)  the definition itself can then be attacked for not adequately defining greenwashing. Still, this is preferable to relying on a broad but vague definition, which would produce unreliable and subjective annotations. Future work should further explore these and other narrow but practical definitions, as they remain useful starting points for uncovering greenwashing practices.

\section{Conclusion}
\label{sec:conclusion}

In this survey, we have provided a comprehensive overview of current research initiatives on detection of greenwashing in textual corporate communications.  
Determining whether a communication contains greenwashing presents three fundamental challenges. First, the absence of a commonly accepted and actionable definition of greenwashing creates ambiguity, complicating any attempt at standardized assessment. Second, there are significant legal and reputational consequences of being accused of greenwashing. Finally, there are no explicit precedents that could serve as jurisprudence for systematic labeling.
This makes the annotation of actual examples of greenwashing challenging, and indeed, there is currently no large-scale dataset of instances of greenwashing.
For this reason, the detection of greenwashing has remained on a theoretical level without empirical validation \citep{moodaley_conceptual_2023}. 
Most existing works target intermediate tasks such as identifying climate-relatedness or green claims. This work also serves its purposes: First, determining whether a text can contain greenwashing at all helps restricting attention to the relevant resources.  
Second, decomposing greenwashing into more manageable steps can lead to an overall better performance, as supported by the principles behind the Chain-of-Thought \citep{cot}. 
Third, using frugal approaches helps reduce costs and energy consumption.
Finally, intermediate tasks can help humans understand the final verdict, which is important for the explainability of decisions. 
Several challenges remain in these works: 
\begin{itemize}
    \item \textit{Evaluation methodology.} As models are capable of tackling more complex tasks, we need tools to correctly evaluate models on these complex tasks. This include building robust evaluation methods for generated texts or building classification benchmarks that are difficult to solve but simple to evaluate. Current results often lack hyper-parameter tuning,  simple baselines, and inter-annotator agreement measures, making it difficult to draw definitive conclusions from them.
    \item  \textit{Model robustness.} When evaluating models, we need to make sure to evaluate also their robustness in real-world applications that exhibit data quality issues or highly imbalanced classes. This includes quantifying the uncertainty of the performance measures. Current works often do not provide confidence intervals or robustness assessments.
    \item \textit{Quantifying information.} Most of the studies we reviewed quantify the amount of text related to a topic, rather than the amount of information. Future work should quantify also the information itself. 
    \item \textit{Relying on regulatory texts.} Most existing works identify one or more supposed characteristics of a misleading claim, but do not confront the statements to regulatory texts. As regulatory texts are becoming more precise about the requirements of environmental communications, they will increasingly become the gold standard %might be sufficient basis 
    for identifying problematic claims.
\end{itemize}

Any step toward understanding the characteristics and formalizing greenwashing is a step toward developing AI systems capable of automatically identifying these misleading communications. 
While previous research has laid a foundation for the theoretical understanding of greenwashing, and for the detection of signals that can indicate misleading climate-related communications, future research should focus on building a representative and diverse corpus of real-world cases of potential greenwashing  to first evaluate models and, secondly, to empirically analyze its mechanisms and patterns. We hope that future research will help identify and combat greenwashing more reliably, thereby fostering transparency and accountability in climate communication.

% \paragraph{Limitations} This survey provides a comprehensive overview of studies on automatic detection of misleading
% climate-related corporate communications and its related tasks as of early 2025. Several limitations remain: Given the extensive scope of the field, our review concentrated on NLP-based research, providing only a limited discussion of topics such as ESG prediction, which are predominantly explored through quantitative methods. Furthermore, our review relied primarily on keyword searches on arXiv, Google Scholar, and ResearchGate, complemented by references from identified articles. In this, we may have overlooked
% relevant studies that use alternative terminology or are published in less accessible venues. Finally, much of greenwashing happens not on the text level, but on the factual level (with claims that contradict data) or on the multimodal level (across images or videos). These are beyond the scope of our survey.

\backmatter

% \bmhead{Supplementary information}

% If your article has accompanying supplementary file/s please state so here. 

% Authors reporting data from electrophoretic gels and blots should supply the full unprocessed scans for key as part of their Supplementary information. This may be requested by the editorial team/s if it is missing.

% Please refer to Journal-level guidance for any specific requirements.

% Acknowledgments
% Acknowledgments of people, grants, funds, etc. should be placed in a separate section on the title page. The names of funding organizations should be written in full.

\bmhead{Acknowledgements}

We are grateful to Takaya Sekine for his insightful feedback on the manuscript. We further acknowledge Amundi’s climate analysts, notably Aaron Macdougall, and the journalists at AEF Info, in particular Nina Godart, for their assistance in researching the topic and refining definitions.

\section*{Declarations}

% Some journals require declarations to be submitted in a standardised format. Please check the Instructions for Authors of the journal to which you are submitting to see if you need to complete this section. If yes, your manuscript must contain the following sections under the heading `Declarations':
% \begin{itemize}
% \item Conflict of interest/Competing interests (check journal-specific guidelines for which heading to use)
% \item Ethics approval and consent to participate
% \item Consent for publication
% \item Materials availability
% \end{itemize}

\subsection*{Funding} This work is funded by Amundi Technology, Inria, Télécom Paris, and ANRT through a CIFRE partnership program.

\subsection*{Conflicts of interest/Competing interests} Aside from the authors’ employers—Amundi Technology, Inria, and Télécom Paris—there are no conflicts of interest related to the present work.

\subsection*{Ethics approval and consent to participate} Not applicable.

\subsection*{Consent for publication} Not applicable.

\subsection*{Data availability} Not applicable.

\subsection*{Materials availability} Not applicable.

\subsection*{Code availability} Not applicable.

\subsection*{Author contribution} This work was primarily conducted by Tom Calamai, particularly the research efforts (collection and analysis of papers). All authors contributed to the writing of the manuscript.

\subsection*{Preprint} An earlier version of this work is available as a preprint on arXiv \citep{calamai2025corporategreenwashingdetectiontext}.

%%%%%%%%%%%%%%%%%% To comment:
\noindent
%If any of the sections are not relevant to your manuscript, please include the heading and write `Not applicable' for that section. 

\bigskip
\begin{flushleft}%

%\bigskip\noindent
%Springer journals and proceedings: \url{https://www.springer.com/gp/editorial-policies}

%\bigskip\noindent
%Nature Portfolio journals: \url{https://www.nature.com/nature-research/editorial-policies}

%\bigskip\noindent
%\textit{Scientific Reports}: \url{https://www.nature.com/srep/journal-policies/editorial-policies}

%\bigskip\noindent
%BMC journals: \url{https://www.biomedcentral.com/getpublished/editorial-policies}
\end{flushleft}

\begin{appendices}

\section{TCFD recommendations}
\label{app:tcfd}
The recommendations are directly taken from the TCFD website: \url{https://www.fsb-tcfd.org/recommendations/}

\paragraph{Governance}Disclose the organization’s governance around climate-related risks and opportunities.
\begin{itemize}
    \item Describe the board’s oversight of climate-related risks and opportunities.
    \item Describe management’s role in assessing and managing climate-related risks and opportunities.
\end{itemize}
 
\paragraph{Strategy} Disclose the actual and potential impacts of climate-related risks and opportunities on the organization’s businesses, strategy, and financial planning where such information is material.
\begin{itemize}
    \item Describe the climate-related risks and opportunities the organization has identified over the short, medium, and long term.
    \item Describe the impact of climate-related risks and opportunities on the organization’s businesses, strategy, and financial planning.
    \item Describe the resilience of the organization’s strategy, taking into consideration different climate-related scenarios, including a 2°C or lower scenario.
\end{itemize}

\paragraph{Risk Management} Disclose how the organization identifies, assesses, and manages climate-related risks.
\begin{itemize}
    \item Describe the organization’s processes for identifying and assessing climate-related risks.
    \item  Describe the organization’s processes for managing climate-related risks.
    \item Describe how processes for identifying, assessing, and managing climate-related risks are integrated into the organization’s overall risk management.
\end{itemize}

\paragraph{Metrics and Targets} Disclose the metrics and targets used to assess and manage relevant climate-related risks and opportunities where such information is material.
\begin{itemize}
    \item Disclose the metrics used by the organization to assess climate-related risks and opportunities in line with its strategy and risk management process.
    \item  Disclose Scope 1, Scope 2 and, if appropriate, Scope 3 greenhouse gas (GHG) emissions and the related risks.
    \item  Describe the targets used by the organization to manage climate-related risks and opportunities and performance against targets.
\end{itemize}

\section{Data sources}
\label{app:data}

\begin{table}[ht]
    \centering
    \begin{tabular}{r|p{8cm}}
    source & papers \\
    \hline
        10-K reports & \citep{doran_risk_disclosure, kolbel_ask_2021, chou_ESG, huangFinBERTLargeLanguage2020, krausEnhancingLargeLanguage2023} \\
        Manually collected reports & \citep{kheradmand2021a, rouenEvolutionESGReports2023, lai_using_2023, hyewon_kang_analyzing_2022, auzepy_evaluating_2023, bjarne_brie_mandatory_2022, nicolas_webersinke_climatebert_2021, luccioni_analyzing_2020, kheradmand2021a, bronzini_glitter_2023, LEE2023119726, marco_polignano_nlp_2022, krausEnhancingLargeLanguage2023} \\ 
        Refinitiv & \citep{SAUTNER_cliamte_change_exp, bingler2023cheaptalkspecificitysentiment, stammbach_environmental_2023, Friederich_climate_risk_disclosure, bjarne_brie_mandatory_2022, LEE_greenwashing, nicolas_webersinke_climatebert_2021} \\
        Reuters & \citep{liCorporateClimateRisk2020, clarkson_nlp_us_csr}\\
        ESG score private providers & \citep{liCorporateClimateRisk2020, SAUTNER_cliamte_change_exp, Mehra_2022, Greenscreen, clarkson_nlp_us_csr, schimanski_bridging_2023} \\
        CDP & \citep{spokoyny2023answering} \\
        Open for Good Data & \citep{avalon_vinella_leveraging_2023} \\
        ClimateWatch & \citep{krausEnhancingLargeLanguage2023} \\
        Twitter/Reddit & \citep{vinicius_woloszyn_towards_2021, luo_detecting_2020, vaid-etal-2022-towards, coanComputerassistedClassificationContrarian2021, divinus_oppong-tawiah_corporate_2023, Greenscreen} \\
        LobbyMap & \citep{morio2023an}\\
        Net-zero tracker & \citep{tobias_schimanski_climatebert-netzero_2023}\\
        Fact-checking websites & \citep{jin-etal-2022-logical, varini_climatext_2020, coanComputerassistedClassificationContrarian2021} \\
        \hline
    \end{tabular}
    \caption{Data Sources}
    \label{tab:data sources}
\end{table}

\paragraph{Publicly available Reports} Many studies rely on 10-K reports~\citep{doran_risk_disclosure, kolbel_ask_2021, chou_ESG, huangFinBERTLargeLanguage2020, krausEnhancingLargeLanguage2023} as they are easily accessible (\href{https://www.sec.gov/edgar}{EDGAR}), uniformly structured, and mandatory for companies listed on US financial markets. Other reports, such as sustainable reports, are not centralized and mandatory, therefore they need to be collected from other sources. They are often collected manually from the company's website~\citep{kheradmand2021a, rouenEvolutionESGReports2023, lai_using_2023, hyewon_kang_analyzing_2022, auzepy_evaluating_2023, bjarne_brie_mandatory_2022, nicolas_webersinke_climatebert_2021}, or from other websites~\citep{luccioni_analyzing_2020, kheradmand2021a, bronzini_glitter_2023, LEE2023119726, marco_polignano_nlp_2022, krausEnhancingLargeLanguage2023} such as from the \href{https://www.globalreporting.org/how-to-use-the-gri-standards/register-your-report/}{GRI}, \href{https://sasb.ifrs.org/company-use/sasb-reporters/}{SASB}, \href{https://www.tcfdhub.org/reports}{TCFD}, \href{https://www.responsibilityreports.com/}{ResponsibilityReports}, \href{https://www.annualreports.com/}{AnnualReports}, \href{https://www.ksa.or.kr}{Korean Standards Association}. \citet{marco_polignano_nlp_2022} relied on Sustainability Disclosure Database from GRI. To the best of our knowledge, it is no longer available. 

The reports are openly available; however, they are most of the time in a PDF format, and must be parsed to produce machine-readable versions. While~\citet{financial-reports-sec} proposed a machine-readable dataset for the 10-K reports, it is often not the case for other types of reports. \citep{bronzini_glitter_2023, marco_polignano_nlp_2022, hyewon_kang_analyzing_2022} provide a sample of processed reports. 

\paragraph{Private Providers} Machine-readable reports, as well as earning call transcripts are often provided through a private data provider such as Refinitiv~\citep{SAUTNER_cliamte_change_exp, bingler2023cheaptalkspecificitysentiment, stammbach_environmental_2023, Friederich_climate_risk_disclosure, bjarne_brie_mandatory_2022, LEE_greenwashing, nicolas_webersinke_climatebert_2021} or Reuters~\citep{liCorporateClimateRisk2020, clarkson_nlp_us_csr}. Similarly, financial data (e.g. ESG score) is often collected through data providers such as RavenPack or S\&P Global Trucost or Wharton’s research platform WRDS, Reuters, Bloomberg, Refinitiv, RobecoSAM and Compustat~\citep{liCorporateClimateRisk2020, SAUTNER_cliamte_change_exp, Mehra_2022, Greenscreen, clarkson_nlp_us_csr, schimanski_bridging_2023}
%Similarly, financial data (e.g. ESG score) is often collected through data providers such as RavenPack \citet{liCorporateClimateRisk2020} or S\&P Global Trucost \citet{SAUTNER_cliamte_change_exp} or Wharton’s research platform WRDS \citep{Mehra_2022, Greenscreen} or Reuters \citet{clarkson_nlp_us_csr}, Bloomberg, Refinitiv, RobecoSAM and Compustat \citep{schimanski_bridging_2023}

\paragraph{Carbon emission data} Carbon emissions of companies can be collected through the private data providers, or through \href{https://www.cdp.net/en/}{CDP} \citep{spokoyny2023answering}, \href{https://www.anderson.ucla.edu/about/centers/impactanderson/open-for-good-transparency-index}{Open for Good Data}~\citep{avalon_vinella_leveraging_2023} or \href{https://www.climatewatchdata.org/}{ClimateWatch website}~\citep{krausEnhancingLargeLanguage2023}.

% \citet{avalon_vinella_leveraging_2023}  for scope 1,2,3 emissions.

\paragraph{Media and Social Media} As companies communicate with the public primarily through marketing and social media, other data sources include Twitter and Reddit~\citep{vinicius_woloszyn_towards_2021, luo_detecting_2020, vaid-etal-2022-towards, coanComputerassistedClassificationContrarian2021, divinus_oppong-tawiah_corporate_2023, Greenscreen}.

%Reddit Climate change dataset contains posts that are related to climate from \href{https://socialgrep.com/datasets/the-reddit-climate-change-dataset}{Socialgrep}.

\paragraph{Climate expert websites} Finally, projects such as \href{https://lobbymap.org/}{LobbyMap} or \href{https://zerotracker.net/}{Net Zero Tracker}, which track corporate policy and engagements, can help build annotated datasets~\citep{morio2023an, tobias_schimanski_climatebert-netzero_2023}. Similarly, relying on a fact-checking website such as \url{https://climatefeedback.org/} or \url{www.skepticalscience.com} is an efficient way of building a dataset~\citep{jin-etal-2022-logical, varini_climatext_2020, coanComputerassistedClassificationContrarian2021}

\section{Reported Performances}\label{app:perf}

% \tom{add in the caption the description of each task (number of labels, ...)}

We have reported all the performances computed in each study when available. We did not include the performance table if the original study did not perform an evaluation (for example, in zero-short/unsupervised settings) or if the evaluation is a heavily modified metric (for example, when using a quantitative measure alongside other signals to compute a financial factor). We only reported traditional NLP metrics. When available, we reported the F1-score, however, if the F1-score is not computed in the study, we reported the available metrics. The computation methods of the F1-score (macro, micro, weighted) are specified when enough information is given to state it confidently. We reported one table per dataset, which might contain multiple tasks, but multiple studies can tackle the same dataset and are then grouped to be able to compare the results. 

\subsection*{Climate-related Topic Classification}

\begin{table*}[ht]
    \centering
    \begin{subtable}[t]{0.5\textwidth}
        \resizebox{\textwidth}{!}{
        \begin{tabular}{rl|ccc}
        \hline
        \textbf{Source} & \textbf{Model} & \textbf{Wikipedia} & \textbf{10-K} & \textbf{Claims}  \\
         & binary & F1 & F1 & F1  \\
        \hline
        \citet{varini_climatext_2020} & best BERT & 80\% $\pm$ 6 & 95 \% $\pm$ 2 & 83\%  $\pm$ 1 \\
        \citet{varini_climatext_2020} & best NB & 60\% $\pm$ 7 & 90 \%  $\pm$ 3 & 72 \%  $\pm$ 2 \\
        \citet{varini_climatext_2020} & best Keywords & 67\%  $\pm$ 7 & 88\% $\pm$ 3 & 70\%  $\pm$ 1\\
        \citet{varini_climatext_2020} & BERT-AL-wiki & 69\%  $\pm$ 6 & 71\% $\pm$ 4 & 81 \%  $\pm$ 1\\
        \citet{garridomerchán2023finetuning} & BERT-AL-wiki & - & 91\% $\pm$ 0.6 & - \\
        \citet{garridomerchán2023finetuning} & climateBERT-AL-wiki & - & 93\% $\pm$ 0.7 & - \\
        \hline
        & & \multicolumn{3}{c}{\textbf{Combined}} \\
        & & \multicolumn{3}{c}{F1 (macro)} \\
        \hline
        \citet{spokoyny2023answering} & Majority & \multicolumn{3}{c}{42.08\%} \\
        \citet{spokoyny2023answering} & Random & \multicolumn{3}{c}{46.86\%} \\
        \citet{spokoyny2023answering} & SVM & \multicolumn{3}{c}{83.39\%} \\
        \citet{spokoyny2023answering} & BERT & \multicolumn{3}{c}{87.04\%} \\
        \citet{spokoyny2023answering} & RoBERTa & \multicolumn{3}{c}{85.97\%} \\
        \citet{spokoyny2023answering} & DistilRoBERTa & \multicolumn{3}{c}{86.06\%} \\
        \citet{spokoyny2023answering} & Longformer & \multicolumn{3}{c}{87.80\%} \\
        \citet{spokoyny2023answering} & SciBERT & \multicolumn{3}{c}{83.29\%} \\
        \citet{spokoyny2023answering} & ClimateBERT & \multicolumn{3}{c}{85.14\%} \\
        \hline
    \end{tabular}
    }
    \caption{Reported performance on ClimaText~\citep{varini_climatext_2020} (annotated datasets).}
    \label{tab:perf_table_climatext}
    \end{subtable}
    \begin{subtable}[t]{0.3\textwidth}
        \resizebox{\textwidth}{!}{
        \begin{tabular}{rl|c}
        \hline
        \textbf{Source} & \textbf{Model} & \textbf{F1 (macro)} \\
        \hline
         \citet{yu_climatebug_2024} & BERT & 90.81\% \\
         \citet{yu_climatebug_2024} & FinBERT & 90.82\% \\
         \citet{yu_climatebug_2024} & ClimateBERT & 91.07\% \\
         \citet{yu_climatebug_2024} & climateBUG-LM & 91.36\% \\
        \hline
    \end{tabular}
    }
    \caption{Reported Performances on ClimateBUG-data~\citet{yu_climatebug_2024}.}
    \label{tab:perf climateBUG}
    \end{subtable}
    \begin{subtable}[t]{0.60\textwidth}
        \resizebox{\textwidth}{!}{
        \begin{tabular}{rl|c}
            \hline
            \textbf{Source} & \textbf{Model} & F1 (weighted average) \\
            \hline
            \citet{nicolas_webersinke_climatebert_2021} & distilRoBERTa & 98.6\% $\pm$ 1 \\
            \citet{nicolas_webersinke_climatebert_2021} & ClimateBERT & 99.1\% $\pm$ 0.8 \\     
            \citet{bingler2023cheaptalkspecificitysentiment} & Naive Bayes & 87\% \\
            \citet{bingler2023cheaptalkspecificitysentiment} & SVM + BoW  & 87\%  \\
            \citet{bingler2023cheaptalkspecificitysentiment} & SVM + ELMo & 89\% \\
            \citet{bingler2023cheaptalkspecificitysentiment} & ClimateBERT & 97\% \\
            \citet{bjarne_brie_mandatory_2022} & ClimateBERT & 98.59\% \\
            \citet{bjarne_brie_mandatory_2022} & FinBERT & 96.67\% \\
            \citet{bjarne_brie_mandatory_2022} & Random Forest & 91.65\% \\  
            \hline
        \end{tabular}
    }
    \caption{Performances on ClimateBERT's climate detection dataset~\citet{bingler2023cheaptalkspecificitysentiment}. \citet{bingler2023cheaptalkspecificitysentiment} also reported an uncertainty measure (p-value) which is not reported in this table for simplicity.}
    \label{tab:perf climatebert climate}
    \end{subtable}
    \caption{Reported performance for the datasets on climate-related topic detection (which include climate, sustainability and environment)}
    \label{tab:reported perf climate}
\end{table*}  

\clearpage

\subsection*{Thematic Analysis}
\label{sec:appendix subtopic}

% \begin{table*}
% \centering

%     \begin{tabular}{ll|cccc}
%     \hline
%     source & model & environment & social & governance & macro f1-score \\ 
%     \hline
%     \citet{schimanski_bridging_2023} & BERT & 91.85 \% & 89.84 \% & 76.86 \% & 86.18 \% \\
%     \citet{schimanski_bridging_2023} & RoBERTa & 92.35 \% & 89.87 \% & 77.03 \% & 86.42 \% \\
%     \citet{schimanski_bridging_2023} & DistilRoBERTa & 90.97 \% & 90.59 \% & 76.65 \% & 86.07 \% \\
%     \citet{schimanski_bridging_2023} & E/S/G-RoBERTa & 93.19 \% & 91.90 \% & 78.48 \% & 87.86 \% \\
%     \citet{schimanski_bridging_2023} & E/S/G-DistilRoBERTa & 92.35 \% & 91.24 \% & 78.86 \% & 87.48 \% \\
%     \end{tabular}
%     \caption{Performance on \citet{schimanski_bridging_2023} with 3 labels}
% \end{table*}

\begin{table*}[ht]
    \centering
    \begin{subtable}[t]{0.65\textwidth}
        \resizebox{\textwidth}{!}{
        \begin{tabular}{rl|cc}
        \hline
        \textbf{Source} & \textbf{Model}      & \textbf{Average P} & \textbf{F1 (macro) } \\ 
        \hline
        \citet{bingler_cheap_2021} & TF-IDF+random forest              & 24 \% & -\\ 
        \citet{bingler_cheap_2021} & Sentence Enc.       & 23\% & -\\ 
        \citet{bingler_cheap_2021} & RoBERTa Para.       & 22\% & -\\ 
        \citet{bingler_cheap_2021} & RoBERTa Sent.       & 75\% & -\\ 
        \citet{bingler_cheap_2021} & ClimateBERT (RoBERTa+LogReg) & 84\% & 84\% \\
        \hline
        \end{tabular}
        }
        \caption{Performance on TCFD recommendations classification \citep{bingler_cheap_2021} with 5 labels}
    \end{subtable}
    \begin{subtable}[t]{0.45\textwidth}
        \resizebox{\textwidth}{!}{
    \begin{tabular}{rl|c}
    \hline
    \textbf{Source} & \textbf{Model} & \textbf{ROC-AUC} \\
    \hline
    \citet{sampson_tcfd-nlp_nodate} & TF-IDF (1) & 0.867 \\
    \citet{sampson_tcfd-nlp_nodate} & kNN (2) & 0.847 \\
    \citet{sampson_tcfd-nlp_nodate} & FT ClimateBERT (3) & 0.852 \\
    \citet{sampson_tcfd-nlp_nodate} & FT distilRoBERTa (4) & 0.819 \\
    \citet{sampson_tcfd-nlp_nodate} & (1) + (2) & 0.884 \\
    \citet{sampson_tcfd-nlp_nodate} & (1) + (3) & 0.865 \\
    \citet{sampson_tcfd-nlp_nodate} & (1) + (4) & 0.857 \\
    \hline
    \end{tabular}
    }
    \caption{Performance on TCFD classification task \citep{sampson_tcfd-nlp_nodate} with 11 labels}
    \end{subtable}
    \begin{subtable}[t]{0.35\textwidth}
        \resizebox{\textwidth}{!}{
        \begin{tabular}{ll|c}
        \hline
        \textbf{Source} & \textbf{Model} & \textbf{F1 (macro)}  \\
        \hline
        \citet{auzepy_evaluating_2023} & BART+MNLI & 56.68\% \\
        \hline
        % source & model &  Micro F1-score: & Macro F1-score: & Weighted F1-score \\
        % \citet{auzepy_evaluating_2023} & BART+MNLI & 60.29 \% & 56.68 \% &   62.81 \% \\
        \end{tabular}
        }
    \caption{Reported performance of the evaluation of the zero-shot approach for TCFD classification (into 11 categories) \citep{auzepy_evaluating_2023}}
    \end{subtable}    
    \caption{Reported performance for the datasets on TFCD classification}
    \label{tab:reported perf tcfd}
\end{table*}    

\begin{table*}[ht]
    \centering
    \begin{subtable}[t]{0.35\textwidth}
    \centering
        \resizebox{\textwidth}{!}{
        \begin{tabular}{rl|c}
        \hline
        \textbf{Source} & \textbf{Model} & \textbf{F1 (micro)} \\ 
        \hline
        \citet{huangFinBERTLargeLanguage2020} & FinBERT        & 89.6\%                    \\
        \citet{huangFinBERTLargeLanguage2020} & BERT           & 86.9\%                   \\
        \citet{huangFinBERTLargeLanguage2020} & NB             & 75.0\%                    \\
        \citet{huangFinBERTLargeLanguage2020} & SVM            & 75.0\%                    \\
        \citet{huangFinBERTLargeLanguage2020} & RF             & 78.3\%                   \\
        \citet{huangFinBERTLargeLanguage2020} & CNN            & 81.9\%                   \\
        \citet{huangFinBERTLargeLanguage2020} & LSTM           & 86.0\%                    \\
        \hline
        \end{tabular}
    }
    \caption{Performance for ESG classification \citep{huangFinBERTLargeLanguage2020}}
    \label{tab:overall_f1_scores_esg_finbert}

        \resizebox{\textwidth}{!}{
    \begin{tabular}{ll|cc}
    \hline
        \textbf{Source} & \textbf{Model} & \textbf{Acc} & \textbf{Stdev}\\
    \hline
        \citet{LEE2023119726} & Multilingual BERT & 84.09 \% & 0.092  \\
        \citet{LEE2023119726} & KoBERT & 85.98 \% & 0.163 \\
        \citet{LEE2023119726} & DistilKoBERT & 84.89 \% & 0.081 \\
        \citet{LEE2023119726} & KB-ALBERT & 85.62 \% & 0.026 \\
        \citet{LEE2023119726} & KLUE-RoBERTa-base & 86.05 \% & 0.125 \\
        \citet{LEE2023119726} & KLUE-RoBERTa-large & 86.66 \% & 0.099 \\
    \hline
    \end{tabular}
    }
    \caption{Performance on ESG classification with 4 labels \citet{LEE2023119726}}
    \end{subtable}
    \begin{subtable}[t]{0.45\textwidth}
        \centering
        \vspace{-50pt}
        \resizebox{\textwidth}{!}{%
        \begin{tabular}{rl|c}
        \hline
        & \textbf{Environment Models}     & \textbf{F1 (unspecified)}           \\ \hline
        \citet{schimanski_bridging_2023} & BERT                            & 91.85\% $\pm$ 1.25                \\
        \citet{schimanski_bridging_2023} & RoBERTa                         & 92.35\% $\pm$ 2.29                \\
        \citet{schimanski_bridging_2023} & DistilRoBERTa                   & 90.97\% $\pm$ 2.00                \\
        \citet{schimanski_bridging_2023} & EnvRoBERTa                      & 93.19\% $\pm$ 1.40                \\
        \citet{schimanski_bridging_2023} & EnvDistilRoBERTa                & 92.35\% $\pm$ 1.65                \\
        & \textbf{Social Models}          &                                  \\ \hline
        \citet{schimanski_bridging_2023} & BERT                            & 89.84\% $\pm$ 0.79                \\
        \citet{schimanski_bridging_2023} & RoBERTa                         & 89.87\% $\pm$ 1.35                \\
        \citet{schimanski_bridging_2023} & DistilRoBERTa                   & 90.59\% $\pm$ 1.03                \\
        \citet{schimanski_bridging_2023} & SocRoBERTa                      & 91.90\% $\pm$ 1.79                \\ 
        \citet{schimanski_bridging_2023} & SocDistilRoBERTa                & 91.24\% $\pm$ 1.86                \\
        & \textbf{Governance Models}      &                                  \\ \hline
        \citet{schimanski_bridging_2023} & BERT                            & 76.86\% $\pm$ 1.77                \\
        \citet{schimanski_bridging_2023} & RoBERTa                         & 77.03\% $\pm$ 1.82                \\
        \citet{schimanski_bridging_2023} & DistilRoBERTa                   & 76.65\% $\pm$ 2.39                \\
        \citet{schimanski_bridging_2023} & GovRoBERTa                      & 78.48\% $\pm$ 2.62                \\
        \citet{schimanski_bridging_2023} & GovDistilRoBERTa                & 78.86\% $\pm$ 1.59                \\
        \hline
        \end{tabular}%
    }%
        \caption{F1-scores (mean $\pm$ std) for environment, social, and governance models under the standard hyperparameter setup.}
        \label{tab:f1_scores_env_social_gov}
    \end{subtable}
    
    \caption{Reported performance for the datasets on ESG classification}
    \label{tab:reported perf esg}
\end{table*}    

\begin{table*}[ht]
    \centering
    \begin{subtable}[t]{0.45\textwidth}
        \resizebox{\textwidth}{!}{
        \begin{tabular}{ll|c}
    \hline
    \textbf{Source} & \textbf{Model} & \textbf{F1 (macro)} \\
    \hline
    \citet{spokoyny2023answering} & Majority & 0.79\% \\
    \citet{spokoyny2023answering} & Random & 5.05\% \\
    \citet{spokoyny2023answering} & SVM & 48.02\% \\
    \citet{spokoyny2023answering} & BERT & 54.74\% \\
    \citet{spokoyny2023answering} & RoBERTa & 52.90\% \\
    \citet{spokoyny2023answering} & DistilRoBERTa & 51.13\% \\
    \citet{spokoyny2023answering} & Longformer & 54.79\% \\
    \citet{spokoyny2023answering} & SciBERT & 51.83\% \\
    \citet{spokoyny2023answering} & ClimateBERT & 52.97\% \\
    \hline
    \end{tabular}%
        }%
    \caption{Performance on SciDCC \citet{mishra2021neuralnere}} \label{tab:SCIDCC perf}
    \end{subtable}
    \begin{subtable}[t]{0.45\textwidth}
        \resizebox{\textwidth}{!}{
        \begin{tabular}{ll|c}
        \hline
        \textbf{Source} & \textbf{Model} & \textbf{F1 (weighted)} \\
        \hline
             \citet{jain_supply_2023} & all-mpnet-base-v2 & 43.7\% \\
             \citet{jain_supply_2023} & TF-IDF & 69\% \\
             \citet{jain_supply_2023} & Word2Vec & 72\% \\
             \citet{jain_supply_2023} & RoBERTa & 87.19\% \\
             \citet{jain_supply_2023} & BERT & 87.14\% \\
             \citet{jain_supply_2023} & ClimateBERT & 85.20\% \\
        \hline
        \end{tabular}
        }
        \caption{Ledger classification \citep{jain_supply_2023}}
        \label{tab:appendix jain}
    \end{subtable}
    \begin{subtable}[t]{0.90\textwidth}
        \resizebox{\textwidth}{!}{
        \begin{tabular}{ll|cccc}
        \hline
        \textbf{Source} & \textbf{Model} & \textbf{Water (\%)} & \textbf{Forest (\%)} & \textbf{Biodiversity (\%)} & \textbf{Nature (\%)} \\ 
        \hline
        \citet{Schimanski2024nature} & EnvironmentalBERT & 94.47 $\pm$ 1.37 & 95.37 $\pm$ 0.92 & 92.76 $\pm$ 1.01 & 94.19 $\pm$ 0.81 \\ 
        \citet{Schimanski2024nature} & ClimateBERT       & 95.10 $\pm$ 1.13 & 95.34 $\pm$ 0.94 & 92.49 $\pm$ 1.03 & 93.50 $\pm$ 0.64 \\ 
        \citet{Schimanski2024nature} & RoBERTa           & 94.55 $\pm$ 0.86 & 94.78 $\pm$ 0.48 & 92.46 $\pm$ 1.54 & 93.97 $\pm$ 0.26 \\ 
        \citet{Schimanski2024nature} & DistilRoBERTa     & 94.98 $\pm$ 1.16 & 95.29 $\pm$ 0.65 & 92.29 $\pm$ 1.23 & 93.55 $\pm$ 0.72 \\ 
        \citet{Schimanski2024nature} & Keywords          & -                & -                & 63.03 & 61.00 \\ 
        \hline
        \end{tabular}
    }
    \caption{Performance (unspecified F1-score) on Nature related topic classification \citep{Schimanski2024nature}}
    \label{tab:appendix nature}
    \end{subtable}
    \begin{subtable}[t]{0.45\textwidth}
        \resizebox{\textwidth}{!}{
        \begin{tabular}{rl|c}
        \hline
         \textbf{Source} & \textbf{Models} & \textbf{F1 (macro)} \\ 
         \hline
        \citet{spokoyny2023answering} & Majority        & 3.65                      \\
        \citet{spokoyny2023answering} & Random          & 6.45                      \\
        \citet{spokoyny2023answering} & SVM             & 58.34                     \\
        \citet{spokoyny2023answering} & BERT            & 64.64         \\
        \citet{spokoyny2023answering} & RoBERTa         & 65.22         \\
        \citet{spokoyny2023answering} & DistilRoBERTa   & 63.61                     \\
        \citet{spokoyny2023answering} & Longformer      & 64.03                     \\
        \citet{spokoyny2023answering} & SciBERT         & 63.62                     \\
        \citet{spokoyny2023answering} & ClimateBERT     & 64.24                     \\
        \hline
        \end{tabular}
        }
        \caption{Performance on the CLIMA-TOPIC classification \citep{spokoyny2023answering}}
        \label{tab:climatopic_performance}
    \end{subtable}
    \begin{subtable}[t]{0.35\textwidth}
         \resizebox{\textwidth}{!}{
            \begin{tabular}{rl|c}
            \hline
            \textbf{Source} & \textbf{Model} & \textbf{F1 (macro)} \\ \hline
            \citet{vaid-etal-2022-towards} & FastText      & 63.8\%            \\
            \citet{vaid-etal-2022-towards} & BERT-Base      & 69.6\%            \\
            \citet{vaid-etal-2022-towards} & BERT-Large    & 69.5\%            \\
            \citet{vaid-etal-2022-towards} & RoBERTa-Base   & 73.4\%            \\
            \citet{vaid-etal-2022-towards} & RoBERTa-Large  & \textbf{73.5\%}   \\
            \citet{vaid-etal-2022-towards} & DistilBERT     & 69.4\%            \\ \hline
            \hline
            \citet{spokoyny2023answering} & Majority        & 13.83\%           \\
            \citet{spokoyny2023answering} & Random         & 16.71\%           \\
            \citet{spokoyny2023answering} & SVM            & 51.81\%           \\
            \citet{spokoyny2023answering} & BERT           & 71.78\%           \\
            \citet{spokoyny2023answering} & RoBERTa        & \textbf{74.58\%}  \\
            \citet{spokoyny2023answering} & DistilRoBERTa  & 72.33\%           \\
            \citet{spokoyny2023answering} & Longformer     & 72.28\%           \\
            \citet{spokoyny2023answering} & SciBERT       & 70.50\%           \\
           \citet{spokoyny2023answering} & ClimateBERT    & 71.83\%           \\ \hline
            \end{tabular}
        }
    \caption{Results for the Fine-grained Classification using \textit{ClimateEng} dataset.}
    \label{tab:performance climateEng}
    \end{subtable}
    \caption{Reported performance for the datasets on climate-related topic detection (other topics)}
    \label{tab:reported perf subtopic}
\end{table*}    

\clearpage

\subsection*{In-depth Disclosure: Climate Risk Classification}

% Intersting: majority is way worse, this might be explained because the majority class is not the same majority class in the test distribution

\begin{table*}[ht]
    \centering
    \begin{subtable}[t]{0.50\textwidth}
        \resizebox{\textwidth}{!}{
        \begin{tabular}{ll|c}
        \hline
            \textbf{Source} &\textbf{Model}           & \textbf{F1 (macro)}   \\ \hline
            \citet{xiang_dare_2023} & BERT-Base                 & 93.1\%         \\
            \citet{xiang_dare_2023} & BERT-Base (Domain pretrained) & \textbf{95.5\%} \\
            \citet{xiang_dare_2023} & TinyBERT                 & 87.0\%         \\
            \citet{xiang_dare_2023} & DistilBERT               & 89.9\%         \\
            \citet{xiang_dare_2023} & ClimateBERT              & 87.5\%         \\
            \citet{xiang_dare_2023} & CCLA+Max-Pooling         & 82.9\%         \\
            \citet{xiang_dare_2023} & Bi-LSTM-Attention        & 71.9\%         \\
            \citet{xiang_dare_2023} & POS-Bi-LSTM-Attention    & 88.2\%         \\
            \citet{xiang_dare_2023} & DARE               & 89.4\%         \\ \hline
        \end{tabular}
        }
    \caption{Performance on sentiment analysis (Risk/Opportunity) \citep{xiang_dare_2023}}
        \label{table:f1_performance dare}
    \end{subtable}
    \begin{subtable}[t]{0.40\textwidth}
        \resizebox{\textwidth}{!}{
       \begin{tabular}{ll|c}
        \hline
         \textbf{Source} & \textbf{Model} & \textbf{F1 (unspecified)} \\
        \hline
        \citet{kolbel_ask_2021} & Bag-of-Words & 83.63\% \\
        \citet{kolbel_ask_2021} & TF-IDF & 80.05\% \\
        \citet{kolbel_ask_2021} & BERT & 94.66\% \\
        \hline
        \end{tabular}
        }
        \caption{Performance for risk classification (General/Transition risk/Physical Risk) \citep{kolbel_ask_2021}}
        \label{tab:f1_scores_cn_tp}
    \end{subtable}
    \begin{subtable}[t]{0.90\textwidth}
        \resizebox{\textwidth}{!}{
        \begin{tabular}{ll|ccccc}
        \hline
        & & \textbf{Binary} &  \textbf{2 Classes} &  & \textbf{5 Classes} & \\
        \textbf{source} & \textbf{Model} & REALISTIC & REALISTIC & REALISTIC & HARD NEG. & DISCRIMINATORY \\
        \hline
        \citet{Friederich_climate_risk_disclosure} & SVM & 29.0\% & 35.1\% & 20.4\% & 45.7\% & 59.9\% \\
        \citet{Friederich_climate_risk_disclosure} & distilBERT & 44.4\% & 49.7\% & 24.1\% & 43.1\% & 55.8\% \\
        \citet{Friederich_climate_risk_disclosure} & RoBERTa & 49.6\% & 44.6\% & 35.6\% & 52.8\% & 59.6\% \\
        \hline
        \end{tabular}
    }
    \caption{Performance (F1  macro average) for climate risk disclosure Tasks from \citet{Friederich_climate_risk_disclosure}}
    \label{tab:performances_friederich}
    \end{subtable}
    \begin{subtable}[t]{0.45\textwidth}
        \resizebox{\textwidth}{!}{
        \begin{tabular}{ll|cc}
    \hline
        \textbf{Source} & \textbf{Model} & \textbf{F1 (unspecified)}\\
    \hline
        \citet{huangFinBERTLargeLanguage2020} & FinBERT & 87.8\% \\
        \citet{huangFinBERTLargeLanguage2020} & BERT & 84.2\% \\
        \citet{huangFinBERTLargeLanguage2020} & NB & 71.1\% \\
        \citet{huangFinBERTLargeLanguage2020} & SVM & 69.6\%  \\
        \citet{huangFinBERTLargeLanguage2020} & RF & 66.8\% \\
        \citet{huangFinBERTLargeLanguage2020} & CNN & 72.5\%  \\
        \citet{huangFinBERTLargeLanguage2020} & LSTM & 73.3\% \\
        \hline
    \end{tabular}
    
        }
        \caption{Performance on sentiment Analysis with FinBERT \citet{huangFinBERTLargeLanguage2020}. }
    \end{subtable}
    \begin{subtable}[t]{0.80\textwidth}
        \resizebox{\textwidth}{!}{
        \begin{tabular}{cc|ccc}
        \hline
            \textbf{Source} & \textbf{Model} & F1 (weighted average) \\
            \hline
            \citet{nicolas_webersinke_climatebert_2021} & distilRoBERTa & 82.5 \% $\pm$ 4.6 \\
            \citet{nicolas_webersinke_climatebert_2021} & climateBERT & 83.8 \% $\pm$ 3.6 \\     
            \citet{bingler2023cheaptalkspecificitysentiment} & Naive Bayes & 72 \% \\
            \citet{bingler2023cheaptalkspecificitysentiment} & SVM + BoW  & 72 \% \\
            \citet{bingler2023cheaptalkspecificitysentiment} & SVM + ELMo  & 75 \%\\
            \citet{bingler2023cheaptalkspecificitysentiment} & climateBERT  & 80 \% \\
            \hline
        \end{tabular}    
        }
        \caption{Performance on risk/opportunity detection task from \citet{nicolas_webersinke_climatebert_2021} and \citet{bingler2023cheaptalkspecificitysentiment}.  \citet{bingler2023cheaptalkspecificitysentiment} also reported an uncertainty measure (p-value) which is not reported in this table for simplicity.}
        \label{tab:perf_table_climatebert_downstream_risk}
    \end{subtable}
    \caption{Reported performance for the datasets on sentiment analysis / risk detection}
    \label{tab:reported perf risk}
\end{table*}  

\clearpage

\subsection*{Green Claim Detection}

\begin{table*}[ht]
    \centering
    \begin{subtable}[t]{0.45\textwidth}
        \resizebox{\textwidth}{!}{
        \begin{tabular}{cc|c}
            \hline
            \textbf{Source} & \textbf{Model} & \textbf{F1 (binary)} \\ 
            \hline
            \citet{stammbach_environmental_2023} & Majority baseline & 00.0\%\\ 
            \citet{stammbach_environmental_2023} & Random baseline & 33.5\%\\ 
            \citet{stammbach_environmental_2023} & ClaimBuster RoBERTa & 33.9\%\\
            \citet{stammbach_environmental_2023} & Pledge Detection RoBERTa & 26.4\%\\
            \hline
            \citet{stammbach_environmental_2023} & TF-IDF SVM & 69.1\%\\
            \citet{stammbach_environmental_2023} & Character n-gram SVM & 70.9\%\\
            \citet{stammbach_environmental_2023} & DistilBERT & 83.7\%\\
            \citet{stammbach_environmental_2023} & ClimateBERT & 83.8\%\\
            \citet{stammbach_environmental_2023} & RoBERTa\_base & 82.4\%\\
            \citet{stammbach_environmental_2023} & RoBERTa\_large & 84.9\%\\
            \hline
        \end{tabular}
        }
        \caption{Performance on environmental claim detection from \citet{stammbach_environmental_2023}}
        \label{tab:perf_env_claim}
    \end{subtable}
    \begin{subtable}[t]{0.45\textwidth}
        \resizebox{\textwidth}{!}{
         \begin{tabular}{cc|c}
            \hline
            \textbf{Source} & \textbf{Model} & \textbf{F1 (binary)} \\ 
            \hline
                \citet{vinicius_woloszyn_towards_2021} & Bin RoBERTa Bal & 92.08\% \\
                \citet{vinicius_woloszyn_towards_2021} & Bin BERTweet Bal & 91.48\% \\
                \citet{vinicius_woloszyn_towards_2021} & Bin Flair Bal & 88.52\% \\
                \citet{vinicius_woloszyn_towards_2021} & Bin RoBERTa Unbal & 88.88\% \\
                \citet{vinicius_woloszyn_towards_2021} & Bin BERTweet Unbal & 84.30\% \\
                \citet{vinicius_woloszyn_towards_2021} & Bin Flair Unbal & 81.55\% \\
            \hline
        \end{tabular}
        }
        \caption{Performance on green claim detection from \citet{vinicius_woloszyn_towards_2021}}
        \label{tab:perf_green_claim}
    \end{subtable}
    \caption{Reported performance for the datasets on claim detection}
    \label{tab:reported perf claims}
\end{table*}

\subsection*{Green Claim Characteristics}

\begin{table*}[ht!]
    \centering
    \begin{subtable}[t]{0.80\textwidth}
        \resizebox{\textwidth}{!}{
        \begin{tabular}{cc|cc}
            \hline
            \textbf{Source} & \textbf{Model} & \textbf{Commitments \& Actions} & \textbf{Specificity}  \\
                     & & F1 (weighted average) & F1 (weighted average) \\ 
            \hline
            \citet{bingler2023cheaptalkspecificitysentiment} & Naive Bayes & 75 \%& 75 \%  \\
            \citet{bingler2023cheaptalkspecificitysentiment} & SVM + BoW   & 76 \%  & 75 \%  \\
            \citet{bingler2023cheaptalkspecificitysentiment} & SVM + ELMo  & 79 \% * & 76 \% \\
            \citet{bingler2023cheaptalkspecificitysentiment} & climateBERT  & 81 \% & 77 \% \\
            \hline
        \end{tabular}
        }
        \caption{Performance on downstream task from \citet{bingler2023cheaptalkspecificitysentiment}. \\citet{bingler2023cheaptalkspecificitysentiment} also reported an uncertainty measure (p-value) which is not reported in this table for simplicity.}
    \label{tab:perf_table_climatebert_downstream}
    \end{subtable}
    \begin{subtable}[t]{0.60\textwidth}
        \resizebox{\textwidth}{!}{
         \begin{tabular}{cc|c}
        \hline
        \textbf{Source} & \textbf{Model} & \textbf{F1(unspecified)} \\ 
        \hline
            \citet{vinicius_woloszyn_towards_2021} & Mult RoBERTa Bal & 81.45 \% \\
            \citet{vinicius_woloszyn_towards_2021} & Mult BERTweet Bal & 75.14 \% \\
            \citet{vinicius_woloszyn_towards_2021} & Mult Flair Bal & 69.96 \% \\
            \citet{vinicius_woloszyn_towards_2021} & Mult RoBERTa Unbal & 76.82 \% \\
            \citet{vinicius_woloszyn_towards_2021} & Mult BERTweet Unbal & 59.71 \% \\
            \citet{vinicius_woloszyn_towards_2021} & Mult Flair Unbal & 38.15 \% \\
        \hline
        \end{tabular}
        }
        \caption{Performance on Implicit/Explicit green claim detection from \citet{vinicius_woloszyn_towards_2021}}
        \label{tab:perf_green_claim_implicit}
    \end{subtable}
    \begin{subtable}[t]{0.60\textwidth}
        \resizebox{\textwidth}{!}{
        \begin{tabular}{ll|cc}
        \toprule
         \textbf{Source} & \textbf{Model} & \textbf{Accuracy} & \textbf{Std} \\
         \hline
            \citet{tobias_schimanski_climatebert-netzero_2023} & ClimateBERT & 96.2 \% & 0.4 \\
            \citet{tobias_schimanski_climatebert-netzero_2023} & DistilRoBERTa & 94.4 \% & 0.7\\
            \citet{tobias_schimanski_climatebert-netzero_2023} & RoBERTa-base & 95.8 \% & 0.6\\
            \citet{tobias_schimanski_climatebert-netzero_2023} & GPT-3.5-turbo & 92.0 \% & - \\
            \bottomrule
        \end{tabular}
        }
        \caption{Performances on net-zero/reduction detection task from \citet{tobias_schimanski_climatebert-netzero_2023}}
        \label{tab:perf_net_zero}
    \end{subtable}
    \caption{Reported performance for the datasets on claim characteristics classification tasks}
    \label{tab:reported perf claims characteristics}
\end{table*}  

\clearpage

 % \begin{subtable}[t]{0.45\textwidth}
 %        \resizebox{\textwidth}{!}{
 %    \begin{tabular}{@{}lcccc@{}}
 %        \toprule
 %        \textbf{Approach} & \textbf{Climate-related} & \textbf{Sentiment} & \textbf{Commitments \& actions} & \textbf{Specificity} \\ 
 %        \midrule
 %        Naive Bayes     & 0.04***                  & 0.05***            & 0.07***                         & 0.08***              \\
 %        SVM + BoW         & 0.05***                  & 0.08***            & 0.11***                         & 0.12***              \\
 %        SVM + ELMo        & 0.03***                  & 0.05***            & 0.06***                         & 0.06***              \\
 %        \bottomrule
 %    \end{tabular}
 %    }
 %    \caption{Evaluation results for cross-validation. This table shows mean improvement of \textsc{ClimateBERT}$_{\text{CTI}}$'s F1 scores over baseline models for different downstream tasks for $n=30$ runs on 800 samples for training and 400 samples for testing. By *, **, and *** we denote $p$-levels below 10\%, 5\%, and 1\%, respectively. \citep{bingler2023cheaptalkspecificitysentiment}}
 %    \label{tab:cross-validation}
 %    \end{subtable}

\subsection*{Green Stance Detection}

\begin{table*}[ht!]
    \centering
    \begin{subtable}[t]{0.45\textwidth}
        \centering
        \vspace{-77pt}
        \resizebox{\textwidth}{!}{
        \begin{tabular}{ll|c}
            \hline
            \textbf{Source} & \textbf{Model} & \textbf{ClimateFEVER} \\
            \hline
            \citet{vaghefi2022deep} & GPT-2 & 67 \%\\ 
            \citet{vaghefi2022deep} & climateGPT-2 & 72 \% \\
            \hline
            \hline
            Claim &  & F1 (macro) \\
            \hline
            \citet{diggelmann_climate-fever_2020} & ALBERT(FEVER) & 32.85\% \\
            \citet{xiang_dare_2023} & BERT & 80.7\% \\
            \citet{xiang_dare_2023} & RoBERTa & 72.3\% \\
            \citet{xiang_dare_2023} & DistilRoBERTa & 72.0\% \\
            \citet{xiang_dare_2023} & ClimateBERT & 76.8\% \\
            \citet{xiang_dare_2023} & POS-Bi-LSTM-Attention & 79.3\% \\
            \citet{nicolas_webersinke_climatebert_2021} & DistilRoBERTa & 74.8\% $\pm$ 3.6\\
            \citet{nicolas_webersinke_climatebert_2021} & ClimateBERT & 75.7\% $\pm$ 4.4 \\
            \hline
            subset 95 examples & & F1 (macro) \\
            \citet{Wang2021EvidenceBA} & RoBERTa Large (FEVER) & 49.2\% \\
            \citet{Wang2021EvidenceBA} & + CLIMATE-FEVER & 64.2\% \\
            \citet{Wang2021EvidenceBA} & + UDA ratio 8 & 71.7\% \\
            \citet{Wang2021EvidenceBA} & + UDA ratio 18 & 71.8\% \\    
            \hline
            \hline
            Evidence &  & F1 (macro) \\
            \hline
            \citet{spokoyny2023answering} & Majority & 26.08\% \\
            \citet{spokoyny2023answering} & Random & 30.62\% \\
            \citet{spokoyny2023answering} & BERT & 62.47\% \\
            \citet{spokoyny2023answering} & RoBERTa & 60.74\% \\
            \citet{spokoyny2023answering} & DistilRoBERTa & 61.54\% \\
            \citet{spokoyny2023answering} & Longformer & 60.82\% \\
            \citet{spokoyny2023answering} & SciBERT & 62.68\% \\
            \citet{spokoyny2023answering} & ClimateBERT & 61.54\% \\        
    
        \end{tabular}
        }
        \caption{Performance on  ClimateFEVER \citet{diggelmann_climate-fever_2020}. The dataset contains pairs of claim and evidences, and an aggregated label for the claim. The evaluation is computed at the claim level (aggregated). Except for \citet{spokoyny2023answering}, they computed the score at the evidence level (for each pair). $\pm$ value reports the standard deviation.}
        \label{tab:perf_table_climatefever}
    
     \resizebox{\textwidth}{!}{
        \begin{tabular}{ll|cc}
            \hline
            \textbf{Source} & \textbf{Model} & \textbf{Acc} & \textbf{F1 (macro)} \\
            \hline
            \citet{luo_detecting_2020} & Majority class & 0.43 & 17\% \\
            \citet{luo_detecting_2020} & Linear & 0.62 & 60\% \\
            \citet{luo_detecting_2020} & BERT & 0.75 & 73\% \\
            \citet{luo_detecting_2020} & Human & 0.71 & - \\
            \hline
        \end{tabular}
    }
    \caption{Performance on GWSD\citep{luo_detecting_2020} (stance detection on global warming)}
    \end{subtable}
    \begin{subtable}[t]{0.4\textwidth}
        \resizebox{\textwidth}{!}{
         \begin{tabular}{ll|c}
            \hline
            \textbf{Source} & \textbf{Model} & \textbf{F1 (macro)} \\
            \hline
            \citet{vaid-etal-2022-towards} & FastText & 34.3\% \\
            \citet{vaid-etal-2022-towards} & BERT-Base & 46.4\% \\
            \citet{vaid-etal-2022-towards} & BERT-Large & 48.9\% \\
            \citet{vaid-etal-2022-towards} & RoBERTa-Base & 51.0\% \\
            \citet{vaid-etal-2022-towards} & RoBERTa-Large & 48.9\% \\
            \citet{vaid-etal-2022-towards} & DistilBERT & 44.8\% \\
            \citet{spokoyny2023answering} & Majority & 29.68\% \\
            \citet{spokoyny2023answering} & Random & 25.52\% \\
            \citet{spokoyny2023answering} & SVM & 42.92\% \\
            \citet{spokoyny2023answering} & BERT & 55.37\% \\
            \citet{spokoyny2023answering} & RoBERTa & 59.69\% \\
            \citet{spokoyny2023answering} & DistilRoBERTa & 52.51\% \\
            \citet{spokoyny2023answering} & Longformer & 34.68\% \\
            \citet{spokoyny2023answering} & SciBERT & 48.67\% \\
            \citet{spokoyny2023answering} & ClimateBERT & 52.84\% \\
            \hline
            \end{tabular}
            
        }
        \caption{Performance on ClimateStance, the stance classification task of \citet{vaid-etal-2022-towards} (3 labels: Favor, Against, Ambiguous)}
            \label{tab:performance ClimateStance}
    
        \resizebox{0.9\textwidth}{!}{
        \begin{tabular}{c|c|c}
        \hline
        \textbf{Model} & \textbf{F1 (macro)} & \textbf{F1 (binary)} \\
        \hline
        \textbf{Baseline Models} & & \\
        BERT & - & 58\% \\
        RoBERTa & - & 89\% \\
        ClimateBERT & 90\% & 90\% \\
        \hline
        \textbf{SeqGAN Approach} & & \\
        BERT & - & 83\% \\
        RoBERTa  & - & 87\% \\
        ClimateBERT & 90\% & 90\% \\
        \hline
        \textbf{MaliGAN Approach} & & \\
        BERT & - & 74\% \\
        RoBERTa & - & 88\% \\
        ClimateBERT & 91\% & 91\% \\
        \hline
        \textbf{RankGAN Approach} & & \\
        BERT & - & 82\% \\
        RoBERTa & - & 87\% \\
        ClimateBERT & 91\% & 90\% \\
        \hline
    \end{tabular}

        }
        \caption{Performance on stance on climate change remediation efforts (support/attack)~\citep{lai_using_2023}}
        \label{tab:appendix-lai_using}
    \end{subtable}
     \begin{subtable}[t]{0.90\textwidth}
        \resizebox{\textwidth}{!}{
            \begin{tabular}{ll|ccc|ccc|ccc}
        \hline
            &  & \multicolumn{3}{c|}{\textbf{Document}} & \multicolumn{3}{c|}{\textbf{Page overlap}} & \multicolumn{3}{c}{\textbf{Strict}} \\
         \textbf{Source}    & \textbf{Model} & P & Q & S & P & Q & S & P & Q & S \\
        \hline
            \citet{morio2023an} & Most-frequent & 46.7\% & 52.6\% & 36.8\% & 51.8\% & 25.6\% & 19.8\% & 41.2\% & 19.6\% & 17.5\% \\
            \citet{morio2023an} & Linear & 66.0 \% & 61.9 \% & 50.3 \% & 71.4 \% & 44.5 \% & 36.1 \% & 52.0 \% & 31.2 \% & 27.0 \% \\
            \citet{morio2023an} & BERT-base & 71.0 \% & 63.5 \% & 51.6 \% & 73.6 \% & 48.1 \% & 37.2 \% & 50.2 \% & 31.9 \% & 25.8 \% \\
            \citet{morio2023an} & ClimateBERT & 71.8 \% & 64.0 \% & 52.8 \% & 74.4 \% & 48.9 \% & 39.0 \% & 50.2 \% & 32.2 \% & 26.8 \% \\
            \citet{morio2023an} & RoBERTa-base & 71.6 \% & 64.5 \% & 53.1 \% & 73.8 \% & 49.6 \% & 38.3 \% & 50.4 \% & 33.4 \% & 26.6 \% \\
            \citet{morio2023an} & Longformer-base & 73.7 \% & 66.9 \% & 54.6 \% & 75.9 \% & 53.0 \% & 40.8 \% & 52.5 \% & 36.1 \% & 28.6 \% \\
            \citet{morio2023an} & Longformer-large & 73.9 \% & 68.8 \% & 57.3 \% & 76.5 \% & 55.0 \% & 44.1 \% & 53.6 \% & 38.7 \% & 31.5 \% \\
        \hline
        \end{tabular}
    }
    \caption{Performance on LobbyMap \citep{morio2023an} using custom metrics described in the study. }
    \label{tab:appendix lobbymap}
    \end{subtable}
    \caption{Reported performance for the datasets on stance detection}
    \label{tab:reported perf stance}
\end{table*}  

% \begin{table*}[ht]
%     \centering
%     \begin{tabular}{ll|c}
%         source & model & F1-score  \\
%         \hline
%         \citet{lai_using_2023} & RoBERTa & 89 \% \\
%         \citet{lai_using_2023} & ClimateBERT & 90 \% \\
%         \citet{lai_using_2023} & BERT & 58 \% \\
%         \citet{lai_using_2023} & RoBERTa + Seq-GAN & 87 \% \\
%         \citet{lai_using_2023} & ClimateBERT + Seq-GAN & 90 \% \\
%         \citet{lai_using_2023} & BERT + Seq-GAN & 83 \% \\
%         \citet{lai_using_2023} & RoBERTa + MALI-GAN & 88 \% \\
%         \citet{lai_using_2023} & ClimateBERT + MALI-GAN & 91 \% \\
%         \citet{lai_using_2023} & BERT + MALI-GAN & 74 \% \\
%         \citet{lai_using_2023} & RoBERTa + RankGAN & 87 \% \\
%         \citet{lai_using_2023} & ClimateBERT + RankGAN & 90 \% \\
%         \citet{lai_using_2023} & BERT + RankGAN & 82 \% \\
%         \hline
%     \end{tabular}
%     \caption{Performance on \citet{lai_using_2023}'s dataset on stance on climate remediation effort}
%     \label{tab:my_label}
% \end{table*}
\clearpage

\subsection*{Question Answering}

\begin{table*}[ht!]
    \centering
    \begin{subtable}[t]{0.70\textwidth}
         \resizebox{\textwidth}{!}{
            \begin{tabular}{l|cccccc}
                \toprule
                \textbf{embeddings} & \textbf{question} & \textbf{definition} & \textbf{concepts} & \textbf{generic} & \textbf{inf\_3} & \textbf{inf\_all} \\
                \hline
                random & 0.037 & 0.037 & 0.037 & 0.037 & 0.037 & 0.037 \\
                BM25 & 0.113 & 0.114 & 0.126 & 0.139 & 0.172 & 0.174 \\
                ColBERTv2 & 0.109 & 0.094 & 0.112 & 0.137 & 0.130 & 0.137 \\
                DRAGON+ & 0.139 & 0.121 & 0.161 & 0.160 & 0.161 & 0.160 \\
                GTE-base & 0.161 & 0.153 & 0.171 & 0.154 & 0.171 & 0.174 \\
                text-embedding-ada-002 & 0.163 & 0.143 & 0.152 & 0.163 & 0.176 & 0.176 \\
                text-embedding-3-small & 0.163 & 0.143 & 0.152 & 0.163 & 0.179 & 0.179 \\
                text-embedding-3-large & 0.167 & 0.143 & 0.152 & 0.163 & 0.179 & 0.179 \\
                \bottomrule
            \end{tabular}
             }
    \caption{Results for the F1-score (custom) of the different embedding models and information retrieval approaches (question, definition, concepts, generic, inf\_3, inf\_all) aggregated across all top-k values (5, 10, 15). \citep{schimanski-etal-2024-climretrieve}}
    \end{subtable}
    \begin{subtable}[t]{0.70\textwidth}
        \resizebox{\textwidth}{!}{
             \begin{tabular}{ll|ccc|ccc}
        \hline
         \textbf{Source} & \textbf{Model} & \multicolumn{3}{c|}{\textbf{Test}} & \multicolumn{3}{c}{\textbf{In-house annotation test}} \\
         & & F1-score & BLEU & METEOR & F1-score & BLEU & METEOR \\
         \hline
         \citet{cliamtebot_2022} & ALBERT & 81.6 \% & 0.678 & 0.808 &  66.1 \% & 0.416 & 0.694 \\
         \hline
    \end{tabular}
    }
    \caption{Performance on CCMRC \citet{cliamtebot_2022}}
    \end{subtable}
    \begin{subtable}[t]{0.45\textwidth}
        \resizebox{\textwidth}{!}{
        \begin{tabular}{ll|cc}
            \hline
            \textbf{Task} & \textbf{Data} & \textbf{Accuracy} & \textbf{(\% of data)} \\ \hline
            \multirow{2}{*}{net zero target year} & raw/optimal & 0.95 & (1.0) \\ 
             & confidence-tuned & 0.97 & (0.5) \\ \hline
            \multirow{3}{*}{reduction target year} & raw & 0.84 & (1.0) \\ 
             & optimal & 0.90 & (0.97) \\ 
             & confidence-tuned & 0.90 & (0.8) \\ \hline
            \multirow{3}{*}{reduction base year} & raw & 0.68 & (1.0) \\ 
             & optimal & 0.80 & (0.84) \\ 
             & confidence-tuned & 0.82 & (0.62) \\ \hline
            \multirow{3}{*}{reduction percentage} & raw & 0.88 & (1.0) \\ 
             & optimal & 0.91 & (0.96) \\ 
             & confidence-tuned & 0.93 & (0.35) \\ \hline
        \end{tabular}
    }
    \caption{Accuracy of the Q\&A approach in assessing the quantitative properties of the net zero and reduction targets~\citep{tobias_schimanski_climatebert-netzero_2023}}
    \label{tab:qa_accuracy}
    \end{subtable}
    \begin{subtable}[t]{0.45\textwidth}
        \resizebox{\textwidth}{!}{
        \begin{tabular}{ll|c}
        \hline
        \textbf{Source} & \textbf{Model} & \textbf{F1 (macro)} \\
        \hline
        \citet{spokoyny2023answering} & Majority & 4.11 \% \\
        \citet{spokoyny2023answering} & Random & 12.14 \% \\
        \citet{spokoyny2023answering} & SVM & 86.00 \% \\
        \citet{spokoyny2023answering} & BERT & 84.57 \% \\
        \citet{spokoyny2023answering} & RoBERTa & 85.61 \% \\
        \citet{spokoyny2023answering} & DistilRoBERTa & 84.38 \% \\
        \citet{spokoyny2023answering} & Longformer & 84.35 \% \\
        \citet{spokoyny2023answering} & SciBERT & 84.43 \% \\
        \citet{spokoyny2023answering} & ClimateBERT & 84.80 \% \\
        \hline
    \end{tabular}
    }
    \caption{Performance on  ClimaINS~\citet{spokoyny2023answering}}
    \label{tab:perf_table_climateins}
    \end{subtable}
    \begin{subtable}[t]{0.45\textwidth}
        \resizebox{\textwidth}{!}{
        \begin{tabular}{l|ccc}
        \hline
         & \textbf{CDP-CITIES} & \textbf{CDP-STATES} & \textbf{CDP-CORP} \\
         \textbf{Model} & \textbf{MRR@10}     & \textbf{MRR@10}     & \textbf{MRR@10}   \\
        \hline
        & \multicolumn{3}{c}{\textbf{No Finetuning on CDP}} \\
        BM25           & 0.055               & 0.084               & 0.153             \\
        MiniLM         & 0.099               & 0.120               & 0.320             \\
        \hline
        & \multicolumn{3}{c}{\textbf{Finetuned on CDP}} \\
                       & \textbf{In-Domain}  & \textbf{In-Domain}  & \textbf{In-Domain} \\
        ClimateBERT    & 0.331               & 0.422               & 0.753             \\
        MiniLM         & 0.366               & 0.482               & 0.755             \\
        \hline
        & \multicolumn{3}{c}{\textbf{Best Model Finetuned on all}} \\
        MiniLM         & 0.352               & 0.489               & 0.745             \\
        \hline
        \end{tabular}
    }
    \caption{MRR@10 scores for BM25, ClimateBERT and MSMARCO-MiniLM on the three subsets of CLIMA-QA. Models finetuned and evaluated on same subset fall under In-Domain.}
    \label{tab:appendix climaqa}
    \end{subtable}
    \begin{subtable}[t]{0.45\textwidth}
        \resizebox{\textwidth}{!}{
            \begin{tabular}{ll|c}
        \hline
        \textbf{Source} & \textbf{Model} & \textbf{F1-score (macro)} \\
        \hline
        \citet{luccioni_analyzing_2020} & RoBERTa-large & 85.5 \% \\
        \citet{luccioni_analyzing_2020} & RoBERTa-base & 82 \% \\
        \hline
    \end{tabular}
    }
    \caption{Performance on question answering \citep{luccioni_analyzing_2020} (2 labels: answer, not) mapping a sentence to one of the TCFD questions}
    \end{subtable}    
    \caption{Reported performance for the datasets on Question/Answering (Retrieval) Tasks}
    \label{tab:reported perf question asnwering (qa)}
\end{table*}  

\clearpage

\begin{table*}[ht]
    \centering
    \begin{subtable}[t]{0.45\textwidth}
        \resizebox{\textwidth}{!}{
        \begin{tabular}{ll|c}
            \hline
            \textbf{Source} & \textbf{Model} & \textbf{Accuracy} \\
            \hline
            \citet{s_vaghefi_chatclimate_2023} & hybrid ChatClimate & 4.38 \\
            \citet{s_vaghefi_chatclimate_2023} & ChatClimate & 4.15 \\
            \citet{s_vaghefi_chatclimate_2023} & GPT-4 & 2.62 \\
            \hline
        \end{tabular}
        }
        \caption{"Accuracy" on a set of 10 questions, human annotations \citet{s_vaghefi_chatclimate_2023}}
        \label{tab:climateIPCC}
    \end{subtable}
    \begin{subtable}[t]{0.4\textwidth}
        \resizebox{\textwidth}{!}{
        \begin{tabular}{ll|c}
            \hline
                \textbf{Source} & \textbf{Model} & \textbf{F1 (macro)} \\
        
            \hline
            \citet{spokoyny2023answering} & Majority & 20.10\% \\
            \citet{spokoyny2023answering} & Random & 24.09\% \\
            \citet{spokoyny2023answering} & SVM & - \% \\
            \citet{spokoyny2023answering} & BERT & 70.57\% \\
            \citet{spokoyny2023answering} & RoBERTa & 71.14\% \\
            \citet{spokoyny2023answering} & DistilRoBERTa & 69.27\% \\
            \citet{spokoyny2023answering} & Longformer & 67.72\% \\
            \citet{spokoyny2023answering} & SciBERT & 68.45\% \\
            \citet{spokoyny2023answering} & ClimateBERT & 69.44\% \\
            \hline        
        \end{tabular}
        }
        \caption{Performance on ClimaBench \citet{spokoyny2023answering}}
        \label{tab:appendix climabench}
    \end{subtable}
    \hspace{\fill}
    \begin{subtable}[t]{0.8\textwidth}
        \centering
        \resizebox{\textwidth}{!}{
        \begin{tabular}{llccccc}
        \hline
        \textbf{Source} & \textbf{Model} & \textbf{ClimaBench} & \textbf{Pira 2.0 MCQ} & \textbf{Exeter Misinf.} & \textbf{Weight. Avg.} \\
        & & F1 (macro) & F1 (unspecified) & F1 (macro) & \\
        \hline
        \citet{thulke2024climategpt} & Stability-3B & 71.4 & 48.7 & 52.6 & 62.8 \\
        \citet{thulke2024climategpt} & Pythia-6.9B & 63.6 & 22.9 & 29.9 & 48.9 \\
        \citet{thulke2024climategpt} & Falcon-7B & 62.9 & 19.8 & 39.9 & 48.3 \\
        \citet{thulke2024climategpt} & Mistral-7B & 73.1 & 19.0 & 63.7 & 53.7 \\
        \citet{thulke2024climategpt} & Llama-2-7B & 68.5 & 51.1 & 59.4 & 62.6 \\
        \citet{thulke2024climategpt} & Jais-13B & 66.9 & 26.4 & 54.2 & 54.4 \\
        \citet{thulke2024climategpt} & Jais-13B-Chat & 65.8 & 66.4 & 61.3 & 64.6 \\
        \hline
        \citet{thulke2024climategpt} & Llama-2-Chat-7B & 67.8 & 72.0 & 64.3 & 68.5 \\
        \citet{thulke2024climategpt} & Llama-2-Chat-13B & 68.6 & 79.3 & 68.6 & 71.4 \\
        \citet{thulke2024climategpt} & Llama-2-Chat-70B & 72.7 & 88.2 & 72.5 & 77.8 \\
        \hline
        \citet{thulke2024climategpt} & ClimateGPT-7B & 75.3 & 86.6 & 65.9 & 77.1 \\
        \citet{thulke2024climategpt} & ClimateGPT-13B & 75.0 & 89.0 & 70.0 & 78.0 \\
        \citet{thulke2024climategpt} & ClimateGPT-70B & 72.4 & 89.9 & 72.5 & 77.7 \\
        \hline
        \citet{thulke2024climategpt} & ClimateGPT-FSC-7B & 59.3 & 17.2 & 45.1 & 46.2 \\
        \citet{thulke2024climategpt} & ClimateGPT-FSG-7B & 53.1 & 17.4 & 41.5 & 42.1 \\
        \hline
        \end{tabular}
        }
        \caption{Results on climate benchmarks\citep{thulke2024climategpt}}
        \label{tab:climate-benchmarks}
    \end{subtable}
    \caption{Reported performance for the datasets on Question/Answering (few-shot and RAG) Tasks.}
     \label{tab:climate QA}
\end{table*}

\clearpage

\subsection*{Deceptive Technique}

\begin{table*}[ht]
    \centering
    \begin{subtable}[t]{0.45\textwidth}
        \resizebox{\textwidth}{!}{
        \begin{tabular}{llr}
            \hline
            \textbf{Source} & \textbf{Model} & \textbf{F1 (micro)} \\
            \hline
            \multicolumn{3}{c}{Trained on general domain} \\
            \hline
            \citet{jin-etal-2022-logical} & Electra-StructureAware & 22.72 \% \\
            \citet{jin-etal-2022-logical} & Electra &  27.23 \% \\
            \multicolumn{3}{c}{Trained on LogicClimate} \\
            \hline
            \citet{jin-etal-2022-logical} & Electra-StructureAware & 23.71 \% \\
            \citet{jin-etal-2022-logical} & Electra & 29.37 \% \\
            \multicolumn{3}{c}{Baseline: Performance on Logic} \\
            \hline
            \citet{jin-etal-2022-logical} & Electra-StructureAware & 58.77 \% \\
            \citet{jin-etal-2022-logical} & Electra & 53.31 \% \\
            \citet{jin-etal-2022-logical} & BERT & 45.80 \% \\
            \hline
        \end{tabular}
        }
        \caption{Performances on \citet{jin-etal-2022-logical} fallacy detection (13 labels)}
        \label{tab:appendix-fallacies}
    \end{subtable}
    \begin{subtable}[t]{0.45\textwidth}
        \resizebox{\textwidth}{!}{
        \begin{tabular}{ll|c}
            \hline
                \textbf{Source} & \textbf{Model} & \textbf{F1 (micro)} \\
                \hline
                \citet{bhatia_automatic_2021-1} & BERT & 59 \% \\
                \citet{bhatia_automatic_2021-1} & BERT* & 62 \% \\
                \citet{bhatia_automatic_2021-1} & MTEXT & 66 \% \\
                \citet{bhatia_automatic_2021-1} & MTEXT* & 67 \% \\
                \citet{bhatia_automatic_2021-1} & MTEXT$_{multi}$ & 68 \% \\
                \citet{bhatia_automatic_2021-1} & MTEXT*$_{multi}$ & 67 \% \\
                \citet{bhatia_automatic_2021-1} & SVM & 49 \% \\
                \citet{bhatia_automatic_2021-1} & Human & 70 \% \\
                \hline
        \end{tabular}
        }
        \caption{Performance on \citet{bhatia_automatic_2021-1}}
        \label{tab:appendix-bhatia-neutralisation}
    \end{subtable}
    \begin{subtable}[t]{\textwidth}
        \centering
        \resizebox{0.45\textwidth}{!}{
            % \begin{tabular}{llccc}
            % \hline
            % Source & Model & F1-Score (\%) \\ \hline
            % \citet{Meddeb2022CounteractingFF} & SVM—6 linguistic features & 58.18 \\
            % \citet{Meddeb2022CounteractingFF} & MNB—bag of words (k = 300) & 77.38 \\
            % \citet{Meddeb2022CounteractingFF} & SVM—bag of words (k = 300) & 76.83 \\
            % \citet{Meddeb2022CounteractingFF} & SVM—bag of words + 6 linguistic features & 77.02 \\ \hline
            % \citet{Meddeb2022CounteractingFF} & BERT on articles & 83.90 \\
            % \citet{Meddeb2022CounteractingFF} & BERT on paragraphs & 84.96 \\
            % \citet{Meddeb2022CounteractingFF} & BERT on sentences & 64.24 \\
            % \citet{Meddeb2022CounteractingFF} & BERT on articles + 6 linguistic features & 84.75 \\ \hline
            % \end{tabular}            
        % }
        % \caption{Results on \citet{Meddeb2022CounteractingFF}}

        \begin{tabular}{ll|c}
    \hline
    \textbf{Source} & \textbf{Model} & \textbf{F1-score (macro)} \\
    \hline
    \citet{coanComputerassistedClassificationContrarian2021} & Logistic (unweighted) &  68 \% \\
    \citet{coanComputerassistedClassificationContrarian2021} & Logistic (weighted) &  72 \% \\
    \citet{coanComputerassistedClassificationContrarian2021} & SVM (unweighted) & 66 \% \\
    \citet{coanComputerassistedClassificationContrarian2021} & SVM (weighted) &  72 \% \\
    \citet{coanComputerassistedClassificationContrarian2021} & ULMFiT & 72 \% \\
    \citet{coanComputerassistedClassificationContrarian2021} & ULMFiT (weighted) & 65 \% \\
    \citet{coanComputerassistedClassificationContrarian2021} & ULMFiT (over sample) &  55 \% \\
    \citet{coanComputerassistedClassificationContrarian2021} & ULMFiT (focal Loss) &  61 \% \\
    \citet{coanComputerassistedClassificationContrarian2021} & ULMFiT-logistic &  75 \% \\
    \citet{coanComputerassistedClassificationContrarian2021} & ULMFiT-SVM & 71 \% \\
    \citet{coanComputerassistedClassificationContrarian2021} & RoBERTa &  77 \% \\
    \citet{coanComputerassistedClassificationContrarian2021} & RoBERTa-logistic & 79 \% \\
    \hline
     & & \textbf{Accuracy (macro)} \\
    \hline
    \citet{coanComputerassistedClassificationContrarian2021} & Coder (Human) &  87 \% \\
    \hline
        & & \textbf{Binary Classification} \\
    \citep{thulke2024climategpt} & Llama-2-Chat-70B & 72.5 \% \\
    \citep{thulke2024climategpt} & ClimateGPT-70B & 72.5 \% \\
    \bottomrule
    \end{tabular}
    }
    \caption{Performance on contrarian claim classification \citep{coanComputerassistedClassificationContrarian2021}}
    \label{tab:perf_contrarian_claim}
    \end{subtable}
    \caption{Deceptive techniques}
\end{table*}

\clearpage

\subsection*{Environmental Performance Prediction}

\begin{table*}[ht]
    \centering
    \begin{subtable}[t]{0.45\textwidth}
        \centering
        \vspace{-25pt}
        \resizebox{\textwidth}{!}{
        \begin{tabular}{ll|c}
        \hline
            \textbf{Source} & \textbf{Model} & \textbf{MSE} \\
        \hline
            \citet{linSUSTAINABLESIGNALSijcai2023} & Baseline & 1.173 \\
            \citet{linSUSTAINABLESIGNALSijcai2023} & Lasso & 0.848 \\
            \citet{linSUSTAINABLESIGNALSijcai2023} & Gradient Boosting & 0.818 \\
            \citet{linSUSTAINABLESIGNALSijcai2023} & NOCATE(RB) & 0.776 \\
            \citet{linSUSTAINABLESIGNALSijcai2023} & NOCATE(CB) & 0.763 \\
            \citet{linSUSTAINABLESIGNALSijcai2023} & NOCATE(DB) &  0.756 \\
            \citet{linSUSTAINABLESIGNALSijcai2023} & SUSTAINABLESIGNALS(RB) & 0.762 \\
            \citet{linSUSTAINABLESIGNALSijcai2023} & SUSTAINABLESIGNALS(CB) & 0.753 \\
            \citet{linSUSTAINABLESIGNALSijcai2023} & SUSTAINABLESIGNALS(DB) & 0.736 \\
        \hline
        \end{tabular}
        }
        \caption{Performance on the task of Finch Score prediction (Finch Score range between 0 and 10) \citet{linSUSTAINABLESIGNALSijcai2023}}
    \end{subtable}
    \begin{subtable}[t]{0.45\textwidth}
    \resizebox{\columnwidth}{!}{
        \begin{tabular}{ll|cc}
        \hline
        \textbf{Source} & \textbf{Model} & \textbf{RMSE} & \textbf{wMAPE} \\
        \hline
            \citet{bronzini_glitter_2023} & OLS & 7.76 & 7.9 \% \\
            Uniform & Random & 40 & 66 \% \\
            Normal & Average & 20 & 31 \% \\
        \hline
        \end{tabular}
        }
        \caption{Performance on ESG score prediction (ESG score range between 0 and 100) \citet{bronzini_glitter_2023}. We added the average RMSE and wMAPE for a uniform distribution of score with a random classifier; and for a normal distribution (mean=50, scale=20) with the average value a prediction.}
        \label{tab:esg bronzini}
    
    \resizebox{\textwidth}{!}{
         \begin{tabular}{ll|cc}
            \hline
            \textbf{Source} & \textbf{Model} & \textbf{F1 (macro)} \\
            \hline
            \citet{clarkson_nlp_us_csr} & Random Forest & 89.48 \% \\
            \citet{clarkson_nlp_us_csr} & XGBoost &  95.73 \% \\
            \hline
        \end{tabular}
        }
        \caption{ \citet{clarkson_nlp_us_csr} (2 classes: good performer, bad performer)}
        \label{tab:perf clarkson_nlp_us_csr}
    \end{subtable}

    \begin{subtable}[t]{0.45\textwidth}
    \centering
    \vspace{-55pt}
    \resizebox{\textwidth}{!}{
         \begin{tabular}{ll|cc}
            \hline
            \textbf{Source} & \textbf{Model} & \textbf{MAE} & \textbf{RMSE} \\
            \hline
            \citet{Greenscreen} & CLIP+MLP & 0.54 & 0.88 \\
            \citet{Greenscreen} & Sentence-BERT+MLP & 0.57 & 0.91 \\
            Uniform & Random & 3.97 & 4.88 \\
            Gaussian & Average & 1.59 & 1.99 \\
            \hline
        \end{tabular}
        }
        \caption{Performance on ESG unmanaged risk score prediction from \citet{Greenscreen} (ESG risk between 0 and 13.5). We added the average RMSE and MAE for a uniform distribution of score with a random classifier; and for a normal distribution (mean=7, scale=2) with the average value a prediction.}
        \label{tab:perf greenscreen}
    \end{subtable}    
    \begin{subtable}[t]{0.45\textwidth}
     \resizebox{\textwidth}{!}{
        \begin{tabular}{c|cp{1.5cm}p{2cm}}
        \hline
            \textbf{Source} & \textbf{Model} & \textbf{Accuracy (Change)} & \textbf{Accuracy (Direction of change)} \\
        \hline
            \citet{Mehra_2022} & majority class & 0.5791 & 0.5682 \\
            \citet{Mehra_2022} & BERT & 0.5985 & 0.4317 \\
            \citet{Mehra_2022} & ESGBERT & 0.6709 & 0.793 \\
        \hline
        \end{tabular}
    }
        \caption{Performance on \citet{Mehra_2022}'s tasks of ESG risk change classification}
        \label{tab:appendix mehra}
    \end{subtable}
    \caption{Environmental Performance Prediction}
    \label{tab:appendix env pred}
\end{table*}

\end{appendices}


\bibliography{sn-bibliography}% common bib file

\end{document}